\documentclass{article} 
\usepackage{iclr2027_conference,times}

\usepackage{amsmath,amsfonts,bm}

\def\eqref#1{equation~\ref{#1}}

\def\1{\bm{1}}

\DeclareMathAlphabet{\mathsfit}{\encodingdefault}{\sfdefault}{m}{sl}
\SetMathAlphabet{\mathsfit}{bold}{\encodingdefault}{\sfdefault}{bx}{n}

\usepackage{hyperref}
\usepackage{url}
\usepackage{booktabs}
\usepackage{multirow}
\usepackage{graphicx}
\usepackage[table]{xcolor}
\usepackage{caption}
\usepackage{amsmath}
\usepackage{algorithmic}
\usepackage{algorithm}
\usepackage{amssymb}
\usepackage{wrapfig}
\usepackage[T1]{fontenc}
\hypersetup{hidelinks}

\title{ADIAS: Automated Design of Interactive Agentic Systems}

\author{Lekang Jiang\thanks{Equal contribution.} \\
University of Cambridge \\
lj408@cam.ac.uk \\
\And
Bohan Tang$^*$ \\
LIGHTSPEED \\
bohantang@global.tencent.com \\
\And
Stephan Goetz \\
University of Cambridge \\
smg84@cam.ac.uk \\
\And
Yiwen Guo\thanks{Corresponding author.} \\
Independent Researcher \\
guoyiwen89@gmail.com
}

\iclrfinalcopy 
\begin{document}

\maketitle

\begin{abstract}

Self-evolving agents can improve without updating model weights by iteratively optimizing the harness based on evaluation feedback. Existing methods are largely candidate-centric: they record \emph{evolution outcomes} (which agent variants have been produced and how they perform), rather than \emph{evolution progress} (which issues, i.e., recurring failure patterns, have been fixed and which remain). Such methods pose two difficulties for cumulative harness evolution. First, repair context is scattered across candidate records and must be reconstructed every round, which leads to inefficient, redundant, or misdirected revisions. Second, because candidates are evaluated as a whole, partial fixes across different agents are difficult to combine, while ineffective changes persist alongside beneficial ones.
To address these problems, we formulate \textbf{issue-centric agent optimization}, which explicitly carries forward evolutionary progress through a persistent issue state.
We instantiate this paradigm in \textbf{ADIAS}, a harness-evolution framework with two key components. An \emph{Issue Manager} maintains the issue state with stable issue identities, lifecycle status, supporting evidence, and intervention-outcome histories. An \emph{Issue-Guided Optimizer} proposes repair targets from the issue state and performs focused harness revision. 
On five interactive benchmarks, ADIAS outperforms the strongest baseline by \textbf{25.2\%} on average and achieves consistent gains across four backbone models. Controlled ablations show that removing the Issue Manager or the Issue-Guided Optimizer reduces performance by up to \textbf{40.7\%}.
\footnote{Code is available at \url{https://github.com/scylj1/adias/}. }

\end{abstract}

\definecolor{best}{RGB}{220,242,220}
\definecolor{second}{RGB}{225,235,250}
\newcommand{\highlightbox}[2]{%
  \begingroup
  \setlength{\fboxsep}{1pt}%
  \colorbox{#1}{\strut #2}%
  \endgroup
}

\section{Introduction}
Large language models (LLMs) are increasingly deployed as \emph{agentic systems}, whose capabilities depend on both backbone models and the surrounding \emph{agent harness} that governs planning, memory, tool use, and other mechanisms \citep{huang2024understanding,wang2024survey}. As model weights often remain fixed after deployment, improving the agent harness offers a practical and efficient way to adapt agents in target environments \citep{weng2026harness,wang2026harness}. However, manually designing effective harnesses is costly. This motivates growing interest in automated agent design or harness self-evolution\footnote{Throughout this paper, \emph{agent design}, \emph{agent optimization}, and \emph{harness evolution} are used interchangeably to refer to modifying the agent harness while keeping the backbone model fixed.}, where an optimizer repeatedly evaluates and revises the agent system itself \citep{maksim2026,gaosurvey2026,yue2026static,ning2026code}.

Agent harness self-evolution generally proceeds through iterative \emph{revise--evaluate--summarize} loops. At each round, the system revises the current agent harness, evaluates the resulting candidate to obtain scores and trajectories, and summarizes these outcomes into a cross-round history for subsequent revision \citep{hu2025automated,zhang2025darwin,lee2026meta,lin2026agentic,chen2026harnessx,zhang2026hyperagents}. Existing methods can be characterized as \textbf{candidate-centric agent optimization}: cross-round experience is organized around candidate agents, and the objective of each round is to propose a better next candidate \citep{lee2026meta, chen2026harnessx}. Such methods 
record \emph{evolution outcomes}: which agents have been produced and how they perform; rather than explicitly tracking \emph{evolution progress}: how issues are addressed and which issues remain unresolved. 

The candidate-centric methods introduce two difficulties for cumulative harness evolution. First, \emph{inefficient optimization targeting}: evidence about whether an issue remains unresolved, which aspect requires modification, and which interventions have already been attempted is scattered across candidate records. The optimizer must reconstruct this repair context before each revision, often resulting in broad, redundant, or misdirected changes. 
Second, \emph{difficult progress consolidation}: candidate-centric methods treat complete candidates as the units of accumulated experience. Consequently, complementary partial repairs to the same issue remain distributed across different candidate branches and are difficult to combine, while ineffective interventions may entangle with beneficial changes and be carried into subsequent rounds. 

\begin{figure*}[t]
    \centering
    \includegraphics[width=.97\textwidth]{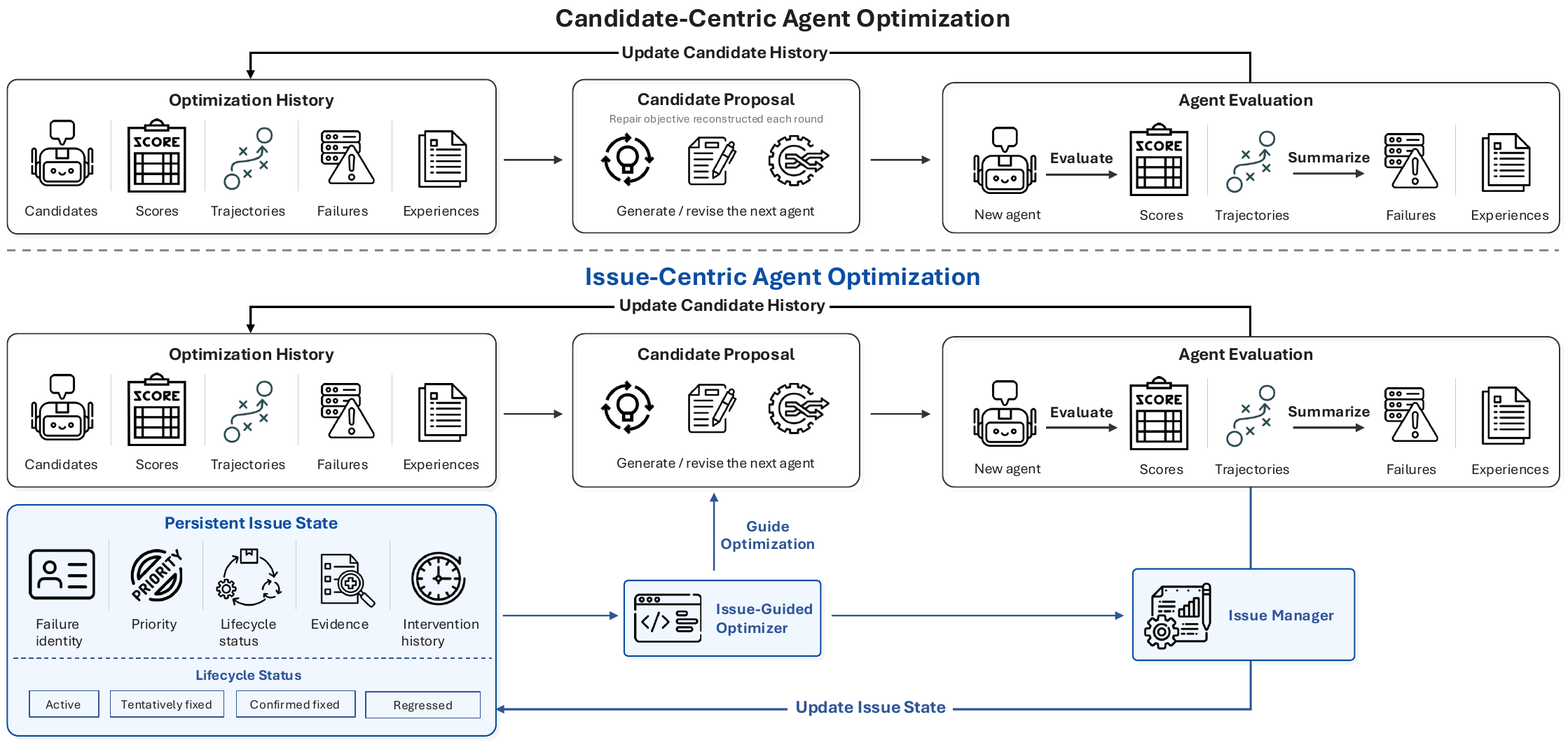}
    \caption{
    Comparison between candidate-centric agent optimization and issue-centric optimization. 
    }
    \label{fig:methodcomp}
\end{figure*}

To fill these gaps, we introduce \textbf{issue-centric agent optimization}, which organizes cross-round experience around persistent issues rather than candidate agents. Evaluation outcomes and diagnostic feedback are accumulated into a persistent issue state that explicitly records each issue's current status, supporting evidence, and intervention history. This state directly guides subsequent repair targets and revision directions, which enables the optimizer to continue, consolidate, or reconsider earlier attempts rather than repeatedly reconstructing the progress from candidate histories. Figure~\ref{fig:methodcomp} contrasts this paradigm with previous candidate-centric evolution.

Realizing issue-centric agent optimization raises two key challenges. First, \emph{reliable issue-state maintenance} requires mapping heterogeneous execution trajectories and sparse outcomes to stable, evidence-grounded issue identities, while reconciling new observations with historical records to track their repair progress across rounds. The same issue may appear differently across candidates, while a single revision may improve some behaviors but regress others, which complicates both identity association and outcome attribution. Second, \emph{actionable and attributable revision} requires translating a dynamic multi-issue state into a concrete code modification. Repair targets and revision directions must be jointly determined from current evidence and intervention history, while modifications must remain focused enough to attribute subsequent outcomes to the intended repair.

To address these challenges, we propose \textbf{ADIAS} (\textbf{A}utomated \textbf{D}esign of \textbf{I}nteractive \textbf{A}gentic \textbf{S}ystems), an issue-centric framework for harness evolution with two key components. First, an \emph{Issue Manager} associates newly diagnosed failures with stable issue identities and tracks their supporting evidence, lifecycle transitions, and intervention--outcome histories. This structure consolidates heterogeneous observations across agent generations while distinguishing the outcome of a targeted repair from simultaneous improvements or regressions in other issues. Second, an \emph{Issue-Guided Optimizer} jointly selects target issues and revision directions from the current state and realizes this plan through focused full-code modification. Joint planning makes the descriptive issue state actionable for agent improvement, while focused revision improves the attribution of subsequent behavioral outcomes without restricting the agent design space. Our main contributions are:

$\bullet$ We propose \textbf{issue-centric agent optimization}, a novel agent harness self-evolution paradigm that organizes cross-round experience around persistent issues and explicitly tracks evolution progress.

$\bullet$ We instantiate this paradigm in \textbf{ADIAS}, a full-code harness-evolution framework with an Issue Manager for tracking evolution progress and an Issue-Guided Optimizer for focused code revision.

$\bullet$ We show that ADIAS outperforms the strongest baseline by \textbf{25.2\%} on average across five interactive benchmarks, with consistent improvements across four backbone models. Controlled ablations illustrate the importance of issue-centric evolution: removing the persistent issue state or replacing issue-guided revision with candidate-centric policies results in performance drops of up to \textbf{40.7\%}.

\section{Related Work}
\label{sec:related_work}

\textbf{Self-Evolving Agent Harness.}
Prior work spans increasing levels of harness expressiveness. Prompt-level methods optimize instructions, demonstrations, or other textual modules while keeping the surrounding agent structure fixed \citep{zhou2022large,pryzant2023automatic,yang2024large,agrawal2026gepa,khattab2024dspy}. Architecture-level methods search over predefined workflows, module compositions, or structured harness primitives \citep{li2024autoflow,zhang2025aflow,zhuge2024gptswarm,shang2025agentsquare,chen2026harnessx}. Full-code methods expose executable agent implementations themselves to optimization, which enables revisions to prompts, memory, planning, tool use, control flow, and other components through iterative search or self-modification \citep{hu2025automated,zhang2025darwin,wang2025huxley,lee2026meta,lou2026autoharness,cai2026moss,nie2026tthe}. ADIAS operates at the full-code level but differs in its search-state control. 

\textbf{Experience and Reusable Knowledge.}
A line of work improves agents by accumulating reusable experience. Reflection and experiential-learning methods retain lessons or heuristics derived from previous trajectories, while systems such as ACE and SkillOpt maintain editable playbooks or skill documents \citep{shinn2023reflexion,zhao2024expel,zhang2025agentic,yang2026skillopt}. Other approaches persist and evolve structured knowledge, workflow, tool, or validation artifacts across iterations \citep{lin2026position,huang2026memoharness,lin2026agentic,wang2026harness}. These methods demonstrate the value of carrying information across rounds. However, the persistent object is primarily reusable knowledge or an editable agent artifact. ADIAS instead maintains the evolving state of the failures that motivate those edits, including their status, evidence, and intervention history.

\section{Method}
\label{sec:method}

\subsection{Task Definition and Candidate-Centric Optimization}
\label{subsec:task-candidate}

\textbf{Task Definition.}
We study the automated design of an interactive agent's harness for a fixed task environment, evaluator, and backbone model, which are isolated from the agent-design workspace. Let $A\in\mathcal{A}$ be an executable agent implementation in the admissible full-code design space $\mathcal{A}$. Executing $A$ on a task instance $x$ yields a trajectory
$
\mathcal{T}(A,x)=(z_1,\ldots,z_L),
$
where
$
z_h=(o_h,a_h,r_h,d_h)
$
records the observation $o_h$ available to the agent, action $a_h$, task score $r_h$, and termination indicator $d_h$ at interaction step $h$. 
We partition task instances into training, validation, and held-out test splits \citep{wang2026rethinking}, and denote aggregate performance on split $q\in{\{\mathrm{train},\mathrm{val},\mathrm{test}\}}$ by $M_q(A)$. Starting from an initial implementation $A_0$ and an optimization budget of $T$ rounds, the design system generates candidate $A_t$ at each round $t=1,\ldots,T$, which is evaluated on task instances to obtain trajectories $\mathcal{T}_t$ and optimization-time performance measurements $M_t$.
After $T$ rounds, the candidate in $\{A_0,\ldots,A_T\}$ with the highest validation performance $M_{\mathrm{val}}(A_t)$ is selected and evaluated on the held-out test set. The objective is to achieve high test performance $M_{\mathrm{test}}$.

\textbf{Candidate-Centric Optimization.}
Most existing automated harness-design methods organize cross-round evolution histories around candidate agents. We formalize such history as $H_t=(h_0,\ldots,h_t), h_k=(A_k,\mathcal{T}_k,M_k,\xi_k), $
where each record $h_k$ is anchored to candidate $A_k$, and $\xi_k$ denotes any method-specific auxiliary experience associated with that candidate, such as reflections, diagnostic summaries, or edit histories. 
These candidate-centric methods can be abstracted as generating the next candidate from candidate-organized history:
\begin{equation}
A_{t+1}
=
\operatorname{Optimizer}(H_t).
\label{eq:candidate_optimizer}
\end{equation}
Here, $\operatorname{Optimizer}$ abstracts over the internal harness evolution procedure rather than assuming a particular revision pipeline. It may reason over previous candidate records and modify an archived candidate agent. 
Although $H_t$ may retain rich behavioral evidence, including explicit failure analysis in $\xi_k$, evidence and interventions concerning the same underlying issue remain distributed across separate candidate records. To determine the next revision, the optimizer must therefore reconstruct the latent issue-level state: which issues remain unresolved, which interventions have been attempted, and what outcomes those interventions produced.

\subsection{Issue-Centric Agent Optimization}
\label{subsec:issue-centric}

Candidate-centric optimization represents cumulative cross-round experience through $H_t$, which leaves the evolution progress implicit and distributed across candidate records. Issue-centric optimization retains $H_t$ but introduces a complementary, persistent issue state $E_t=\{e_i^t\}_{i\in\mathcal{I}_{\leq t}}, $ where $\mathcal{I}_{\leq t}$ denotes the set of issues identified up to round $t$, and $e_i^t$ records the accumulated evolution progress associated with issue $i$. Whereas each record $h_k\in H_t$ is indexed by a candidate at a particular round, each record $e_i^t\in E_t$ is indexed by an underlying issue and persists across candidate generations. Evidence, attempted interventions, and their outcomes can therefore be accumulated around the same issue rather than remaining attached to separate candidates. At each round, issue-centric optimization updates the issue state and conditions the next revision on both representations:
\begin{align}
E_t
&=
\operatorname{IssueUpdate}(E_{t-1},H_t),
\label{eq:generic_issue_update}\\
A_{t+1}
&=
\operatorname{Optimizer}(H_t,E_t).
\label{eq:issue_centric_optimizer}
\end{align}
Here, $\operatorname{IssueUpdate}$ abstracts any procedure that identifies new issues, associates new evidence with existing issues, and consolidates outcomes of attempted interventions. The optimizer then uses both $H_t$ and $E_t$ to determine subsequent revisions. Therefore, while candidate-centric optimization needs to repeatedly reconstruct issue-level progress from $H_t$, issue-centric optimization explicitly maintains and carries this progress across rounds.

\begin{figure*}[t]
    \centering
    \includegraphics[width=\textwidth]{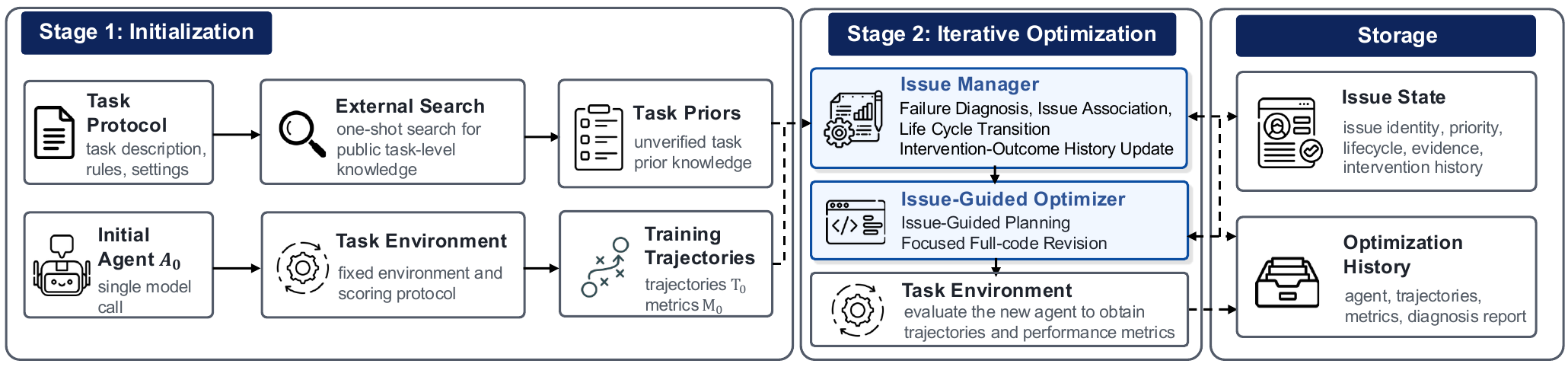}
    \caption{
     Overview of ADIAS that consists of two stages: initialization and iterative optimization.
    }
    \label{fig:method}
\end{figure*}

\subsection{ADIAS}
\label{sec:adias}

ADIAS instantiates the issue-centric formulation in Eq.(\ref{eq:generic_issue_update}) and Eq.(\ref{eq:issue_centric_optimizer}) with two concrete components: an \emph{Issue Manager} and an \emph{Issue-Guided Optimizer}. The Issue Manager maintains a persistent issue state, while the Issue-Guided Optimizer achieves issue-conditioned harness code modification. Figure~\ref{fig:method} provides an overview workflow of ADIAS, which consists of two stages: initialization and iterative optimization. The \textbf{initialization stage} searches for task-level priors and behavioral evidence collected from the initial agent. ADIAS begins with an initial agent $A_0$ with a single model call. To reduce cold-start uncertainty, a one-shot search over publicly available task-level information produces provisional priors $P$ about task requirements, likely failure modes, and potentially useful agent mechanisms. $P$ is incorporated into the initial issue state as unverified task prior knowledge. Details are in Appendix~\ref{app:external-prior}. The environment evaluates $A_0$ to obtain initial trajectories $\mathcal{T}_0$ and performance metrics $M_0$. 
The \textbf{optimization stage} then proceeds as a three-step loop: Issue Manager updates the issue state, Issue-Guided Optimizer modifies the agent code, and environment evaluates the new agent to get trajectories and metrics to update the optimization history. 

\subsubsection{Issue Manager}
\label{sec:manager}

The issue record $e_i^t$ is designed as $e_i^t = (id_i, q_i^t, s_i^t, \mathcal{B}_i^t, \mathcal{U}_i^t), $ where $id_i$ is a stable identity, $q_i^t$ is priority, $s_i^t$ is lifecycle status, $\mathcal{B}_i^t$ is supporting evidence, and $\mathcal{U}_i^t$ is intervention-outcome history. The Issue Manager maintains this state by failure diagnosis, issue association, lifecycle transition, and intervention-outcome history update.

\textbf{Failure Diagnosis.}
At each round $t \ge 1$, a dedicated diagnostic agent analyzes trajectories $\mathcal{T}_{t-1}$ and performance metrics $M_{t-1}$ of agent $A_{t-1}$, and produces a diagnostic report $D_t$ containing observed failure categories, supporting evidence (trajectories that support the failure), and inferred causes. Details are in Appendix~\ref{app:diagnosis}.
This process converts candidate-level execution evidence into observations that can be reconciled with the persistent issue state.

\textbf{Issue Association.}
For each issue in $D_t$, the Issue Manager either associates it with an existing issue in $E_{t-1}$ or creates a new persistent identity $id_i$ based on the failure category. Repeated observations and supporting evidence $\mathcal{B}_i^t$ are merged into the corresponding issue record, which preserves issue identity across rounds and allows evolution progress to accumulate without rediscovering the same underlying failure. Each issue is also assigned a priority $q_i^t$ based on issue severity, as determined by LLMs. Details of issue control are in Appendix~\ref{subsec:issuecontrol}.

\textbf{Lifecycle Transition.}
The issue status $s_i^t$ transitions among \textsc{active}, \textsc{tentatively-fixed}, \textsc{confirmed-fixed}, and \textsc{regressed}. New and currently observed issues are marked \textsc{active}, while a previously fixed issue that reappears is marked \textsc{regressed}. An \textsc{active} or \textsc{regressed} issue becomes \textsc{tentatively-fixed} once absent, and becomes \textsc{confirmed-fixed} after remaining absent for $\alpha_{\min}$ consecutive evaluations ($\alpha_{\min}=2$ used in our experiments). These transitions distinguish temporary disappearance from reliable resolution while making recurring failures explicit.

\textbf{Intervention-Outcome History Update.}
For each targeted issue, the Issue Manager records the round-$t$ revision and its outcome in $\mathcal{U}_i^t$. The targeted intervention outcome is stored separately from lifecycle changes observed for other issues, making progress on the intended repair distinguishable from concurrent improvements or regressions elsewhere. Supported intervention outcomes are distilled into reusable repair lessons. This issue-specific history supports outcome attribution and helps subsequent rounds reuse effective interventions while avoiding previously unsuccessful directions.

\subsubsection{Issue-Guided Optimizer}
\label{sec:reviser}

Issue-Guided Optimizer comprises two steps: issue-guided planning and focused full-code revision.

\textbf{Issue-Guided Planning.}
At round $t$, an LLM optimizer jointly determines a parent candidate and revision plan as $(A_{p_t},R_t)=\operatorname{Plan}(H_{t},E_{t})$, specifying \emph{what to repair}, \emph{where to resume optimization}, and \emph{how to revise}. It first selects a small set of target issues from $E_{t}$, prioritizing issues that are severe, repeatedly observed, currently active, or recently regressed. It then selects $A_{p_t}$ from $H_{t}$ based on its repair history with respect to the target issues, which favors candidates that preserve useful progress on these issues to provide a suitable starting point for further evolution. Finally, the optimizer analyzes previous intervention outcomes to specify corresponding revision plan in $R_t$.

\textbf{Focused Full-Code Revision.}
The optimizer executes the plan as $A_{t}=\operatorname{Modify}(A_{p_t},R_t)$ under two fixed constraints. First, the entire full-code space $\mathcal{A}$ remains editable, which allows revisions to any agent component, including prompts, observation processing, memory, planning, tool-use policies, control flow, verification, and recovery mechanisms. Second, each revision modifies at most $K$ harness components ($K=3$ in our experiments), which keeps the intervention focused for more reliable outcome attribution while preserving full-code flexibility.

The resulting agent $A_t$ is evaluated to obtain $\mathcal{T}_t$ and $M_t$. In the following round, its execution behavior is diagnosed to produce the diagnostic report $D_{t+1}$, after which the optimization history and issue state are updated to $H_{t+1}$ and $E_{t+1}$, respectively. The updated pair $(H_{t+1},E_{t+1})$ then guides the next optimization round.

\section{Experiments}
\label{sec:experiments}

\subsection{Experimental Setup}

\textbf{Benchmarks.}
We evaluate ADIAS on five interactive agent benchmarks spanning different scenarios, including $\tau$-Bench for tool-use in customer service \citep{yao2025tau}, ALFWorld for embodied AI planning \citep{shridharalfworld}, TextCraft for compositional crafting games \citep{prasad2024adapt}, WebShop for web navigation \citep{yao2022webshop}, and ScienceWorld for scientific experimentation \citep{wang2022scienceworld}. These environments require agents to make a sequence of state-dependent decisions rather than produce a single static response. To control computational costs, we construct representative task subsets for the main experiments. We report benchmark details in Appendix \ref{app:benchmarks}. 

\textbf{Baselines.}
We compare ADIAS with five representative baselines on different editable scopes and experience organization. We use a manually designed ReAct-style agent \citep{yao2023react} with task-level memory as the fixed, non-automated baseline. SkillOpt \citep{yang2026skillopt} performs prompt-level agent optimization by maintaining a skill document. Agentic Harness Engineering (AHE) \citep{lin2026agentic} performs architecture-level optimization over pre-defined components of agent harnesses. Meta-Harness \citep{lee2026meta} performs end-to-end optimization over the entire agent harness. DGM-Hyperagents (DGM-H) \citep{zhang2026hyperagents} integrates the task agent and meta agent into a single editable program and allows the resulting hyperagent to modify both its task-solving logic and the mechanism used to generate future improvements. All methods use the same benchmark wrappers, task splits, action interfaces, and scoring scripts. Details are in Appendix~\ref{app:baselines}.

\textbf{Models.}
We use DeepSeek-V4-Flash \citep{deepseek2026v4} as the default backbone model. To evaluate cross-model robustness, we additionally conduct experiments with GLM-5.2 \citep{zai2026glm52}, Hy3-Preview \citep{tencent2026hy3}, and GPT-5.4 \citep{openai2026gpt54}. For each experimental setting, the same backbone model is used for both the design agent and the task agent. We apply this matched-model protocol to all automated baselines and ADIAS, which ensures that performance differences are attributable to the agent-design method rather than to a stronger optimizer model. For $\tau$-Bench, the user simulator is held fixed to DeepSeek-V4-Flash across all settings.

\textbf{Evaluation.}
For each method, we select the agent with the highest validation performance and evaluate it on the held-out test set. All methods use the same task splits, environment interfaces, optimization budgets, and benchmark-specific evaluators. We report the native metric of each benchmark (\%): success rate for $\tau$-Bench, ALFWorld and TextCraft, and the normalized native task score for WebShop and ScienceWorld. No test tasks or validation trajectory feedback are exposed during optimization. Moreover, we report interaction efficiency, which measures task performance per environment interaction. Since the semantics of an interaction step differ across environments, efficiency is compared only within the same benchmark. More details are in Appendix~\ref{app:evaluation-protocol}.

\textbf{Experimental Settings.}
We run each automated method for a fixed budget of 10 optimization iterations. At each iteration, 15 training episodes are sampled for agent optimization. The same iteration and rollout budgets are applied to all methods within each benchmark. Each task-agent episode is limited to 30 environment interaction steps or dialogue turns. We also keep decoding and inference configurations fixed across methods using the same backbone model. We separately report meta-level optimization token consumption and controlled cost comparisons in Appendix~\ref{app:token_cost}.

\subsection{Results}

\subsubsection{Main Results}

We report task performance, cross-model robustness, and optimization efficiency in the main text. Additional analyses of execution robustness, optimization variability, test-set uncertainty, and broader task distributions are provided in Appendix~\ref{app:moreresults}.

\textbf{Strong Task Performance and Efficiency.}
Table~\ref{tab:main_results} compares ADIAS with fixed handcrafted strategies and recent automated design baselines across five interactive benchmarks. ADIAS achieves the best task performance and interaction efficiency on all five benchmarks, which shows robust generalization across environments with diverse interaction patterns and task objectives. Averaged across benchmarks, ADIAS reaches a score of 78.4 versus 62.6 for the strongest baseline (DGM-H), a relative improvement of 25.2\%. Specifically, scores of ADIAS on $\tau$-Bench, ALFWorld, TextCraft, WebShop, and ScienceWorld are 81.3, 94.0, 91.0, 69.4, and 56.3, respectively, which consistently surpass all baselines.
These gains are not obtained through longer interactions. ADIAS also achieves the highest interaction efficiency on every benchmark, which indicates that its learned strategies translate environment interactions into task progress more effectively. The advantage is particularly pronounced on ALFWorld and TextCraft, where ADIAS reaches efficiencies of 8.95 and 9.01, and substantially exceeds the strongest baseline results of 4.48 (AHE) and 5.98 (DGM-H), respectively.

\begin{table*}[t]
\centering
\caption{
Task performance and efficiency across five interactive benchmarks using DeepSeek-V4-Flash. Best and second-best results are highlighted in \highlightbox{best}{green} and \highlightbox{second}{blue}, respectively.
}
\label{tab:main_results}
\small
\resizebox{\textwidth}{!}{
\begin{tabular}{lcc cc cc cc cc c}
\toprule
\multirow{2.5}{*}{Method}
& \multicolumn{2}{c}{$\tau$-Bench-Return}
& \multicolumn{2}{c}{ALFWorld}
& \multicolumn{2}{c}{TextCraft}
& \multicolumn{2}{c}{WebShop}
& \multicolumn{2}{c}{ScienceWorld}
& Avg. \\
\cmidrule(lr){2-3}
\cmidrule(lr){4-5}
\cmidrule(lr){6-7}
\cmidrule(lr){8-9}
\cmidrule(lr){10-11}
& Score $\uparrow$ & Eff. $\uparrow$
& Score $\uparrow$ & Eff. $\uparrow$
& Score $\uparrow$ & Eff. $\uparrow$
& Score $\uparrow$ & Eff. $\uparrow$
& Score $\uparrow$ & Eff. $\uparrow$
& Score $\uparrow$ \\
\midrule

Handcrafted
& \cellcolor{second}75.0
& \cellcolor{second}5.73
& 70.9
& 3.60
& 38.0
& 1.52
& 39.1
& 2.13
& 38.1
& 3.05
& 52.2 \\

SkillOpt
& 12.5
& 0.45
& 0.0
& 0.00
& 37.0
& 1.70
& 22.3
& 1.17
& 30.1
& 1.89
& 20.4 \\

Meta-Harness
& 43.8
& 2.39
& 59.7
& 3.32
& 24.0
& 0.95
& 46.0
& 3.71
& 26.5
& 1.65
& 40.0 \\

AHE
& 56.2
& 3.05
& \cellcolor{second}78.4
& \cellcolor{second}4.48
& 35.0
& 1.39
& 57.3
& \cellcolor{second}5.46
& 42.6
& 3.44
& 53.9 \\

DGM-H
& 59.4
& 3.93
& 69.4
& 3.42
& \cellcolor{second}76.0
& \cellcolor{second}5.98
& \cellcolor{second}57.5
& 5.42
& \cellcolor{second}50.9
& \cellcolor{second}3.98
& \cellcolor{second}62.6 \\

ADIAS (Ours)
& \cellcolor{best}81.3
& \cellcolor{best}6.35
& \cellcolor{best}94.0
& \cellcolor{best}8.95
& \cellcolor{best}91.0
& \cellcolor{best}9.01
& \cellcolor{best}69.4
& \cellcolor{best}5.55
& \cellcolor{best}56.3
& \cellcolor{best}3.99
& \cellcolor{best}78.4 \\

\bottomrule
\end{tabular}
}
\end{table*}

\begin{table*}[t]
\centering
\small
\caption{
Cross-model task performance and efficiency on $\tau$-Bench-Return. Best and second-best results are highlighted in \highlightbox{best}{green} and \highlightbox{second}{blue}, respectively.
}
\label{tab:cross_model}
\resizebox{\textwidth}{!}{
\begin{tabular}{lcc cc cc cc c}
\toprule
\multirow{2.5}{*}{Method}
& \multicolumn{2}{c}{DeepSeek-V4-Flash}
& \multicolumn{2}{c}{GLM-5.2}
& \multicolumn{2}{c}{Hy3-Preview}
& \multicolumn{2}{c}{GPT-5.4}
& \multirow{2.5}{*}{Avg. Score $\uparrow$} \\
\cmidrule(lr){2-3}
\cmidrule(lr){4-5}
\cmidrule(lr){6-7}
\cmidrule(lr){8-9}
& Score $\uparrow$ & Eff. $\uparrow$
& Score $\uparrow$ & Eff. $\uparrow$
& Score $\uparrow$ & Eff. $\uparrow$
& Score $\uparrow$ & Eff. $\uparrow$
& \\
\midrule

Handcrafted
& \cellcolor{second}75.0
& \cellcolor{second}5.73
& \cellcolor{second}84.4
& \cellcolor{second}6.49
& 15.6
& 0.86
& 71.9
& \cellcolor{second}5.89
& 61.73 \\

SkillOpt
& 12.5
& 0.45
& 12.5
& 0.67
& 12.5
& 0.59
& 15.6
& 0.74
& 13.28 \\

Meta-Harness
& 43.8
& 2.39
& 78.1
& 6.51
& 68.8
& 5.38
& 65.6
& 5.56
& 64.08 \\

AHE
& 56.2
& 3.05
& 62.5
& 4.08
& \cellcolor{second}78.1
& \cellcolor{second}6.20
& \cellcolor{second}75.0
& 5.81
& 67.95 \\

DGM-H
& 59.4
& 3.93
& 65.6
& 4.10
& 75.0
& 5.86
& 71.9
& 5.33
& \cellcolor{second}67.98 \\

ADIAS (Ours)
& \cellcolor{best}81.3
& \cellcolor{best}6.35
& \cellcolor{best}90.6
& \cellcolor{best}7.13
& \cellcolor{best}84.4
& \cellcolor{best}6.54
& \cellcolor{best}87.5
& \cellcolor{best}6.63
& \cellcolor{best}85.95 \\

\bottomrule
\end{tabular}
}
\end{table*}

\textbf{Cross-Model Robustness.} We conduct a cross-model evaluation on $\tau$-Bench-Return using four different models: DeepSeek-V4-Flash, GLM-5.2, Hy3-Preview, and GPT-5.4. Table~\ref{tab:cross_model} shows that ADIAS consistently achieves the highest performance across all backbone models. The corresponding interaction efficiencies are also consistently the highest among all compared methods. These results show that ADIAS provides robust harness evolution across different backbone models.

\begin{wrapfigure}{r}{0.45\columnwidth}
    \centering
    \includegraphics[width=\linewidth]{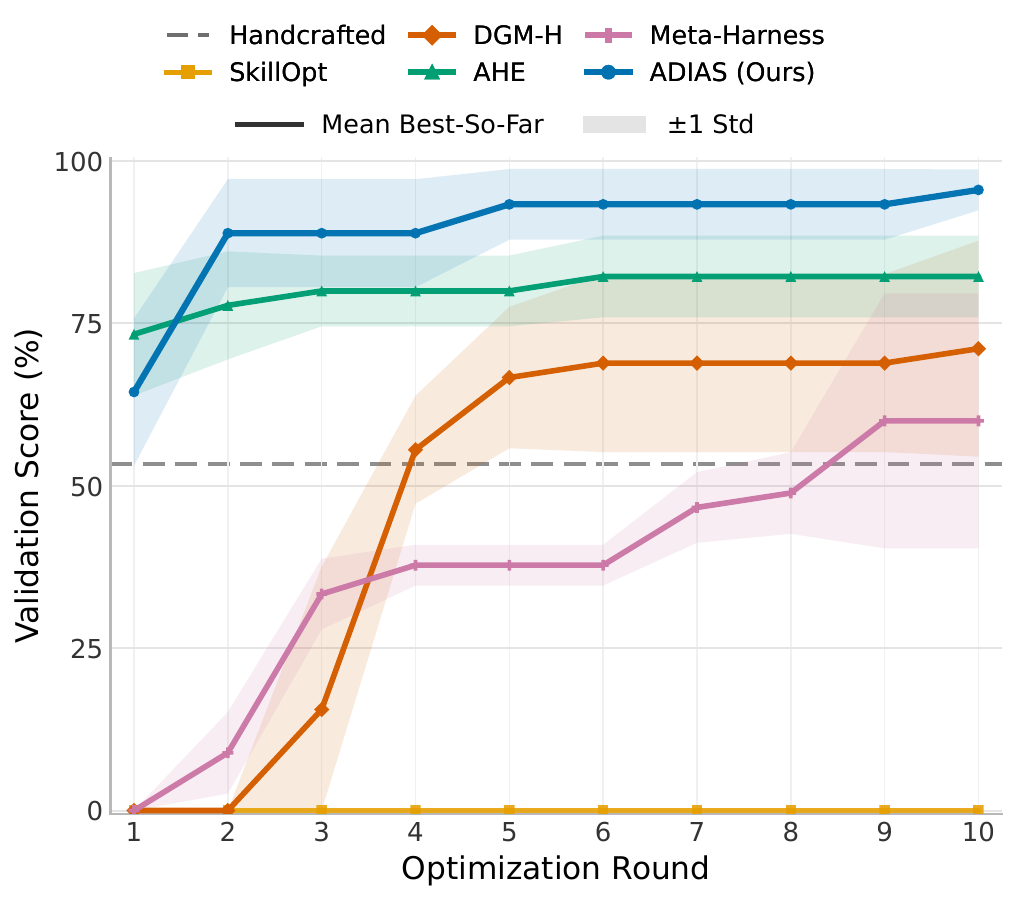}
    \caption{Results on ALFWorld.}
    \label{fig:optimization_progress}
\end{wrapfigure}

\textbf{High Optimization Efficiency and Stability.}
Figure~\ref{fig:optimization_progress} reports the mean best-so-far validation performance across three independent ALFWorld runs, with shaded regions indicating $\pm1$ standard deviation. ADIAS rapidly discovers strong agent designs, reaching nearly 90\% validation performance by the second round and maintaining the highest mean best-so-far score throughout the remaining optimization process. Its consistently strong performance across independent runs indicates that the gains are not specific to a single favorable optimization trajectory. In comparison, DGM-H and Meta-Harness require more rounds to reach competitive performance. SkillOpt shows limited progress as it modifies only a persistent textual skill document and cannot introduce additional agent components, such as explicit memory mechanisms, needed for effective interaction in ALFWorld.

\subsubsection{Ablation Study}
\label{subsec:ablation}

As shown in Table~\ref{tab:ablation}, we ablate three coupled aspects of our proposed Issue Manager and the Issue-Guided Optimizer: the evidence used to update issue states, the issue-level representation for cross-round persistence, and whether this state actively guides optimization.

\textbf{External priors and round-level diagnosis provide necessary evidence.} 
Removing either source reduces both performance and interaction efficiency across all three benchmarks. Without the external prior, the largest drop occurs on TextCraft, from 91.0 to 61.0. Without round-level diagnosis, ALFWorld decreases from 94.0 to 63.4. These results indicate that the two mechanisms play important roles: the external prior reduces cold-start uncertainty, while trajectory diagnosis provides the behavioral evidence required to ground and update the issue state. 

\textbf{A raw candidate archive cannot replace persistent issue state.}
\emph{Archive-Wide Synthesis} removes the durable issue state and follows a Meta-Harness-style mechanism \citep{lee2026meta}. The designer receives a raw archive containing candidate agents, evaluation results, trajectory histories, and diagnostic reports, and synthesizes the next optimization step directly from this archive. Its scores decrease to 65.6, 60.4, and 32.0 on $\tau$-Bench-Return, ALFWorld, and TextCraft, respectively. Although the archive preserves rich historical evidence, identifying an effective revision direction remains difficult in interactive environments with long trajectories and sparse feedback, where useful signals must be recovered from heterogeneous candidate-level records.

\textbf{Maintaining issue state is insufficient unless it controls optimization.}
\emph{Best-Candidate Revision} and \emph{Latest-Candidate Continuation} retain the same candidate history and persistent issue states as ADIAS, but the issue state is treated only as contextual information: the actual parent is selected by validation score or generation order, rather than the Issue-Guided Optimizer.
\emph{Best-Candidate Revision}, the strongest ablation, still underperforms ADIAS by 9.4, 18.6, and 17.0 points across the three benchmarks. As each round returns to the best-score candidate, locally useful interventions discovered in different branches remain isolated and are difficult to consolidate into a single improved lineage.
\emph{Latest-Candidate Continuation} exhibits another limitation. As every round continues from the most recent candidate, the next revision is more likely to inherit ineffective or regressive modifications. The parents selected by these two methods are not necessarily the most suitable starting point for the current evolution. The persistent issue state should therefore serve as an operational control signal to guide optimization.

\begin{table*}[t]
\centering
\caption{
Ablation study on representative benchmarks using DeepSeek-V4-Flash. Average Score is the unweighted average across the three benchmarks, and $\Delta$ denotes its change relative to full ADIAS. Best and second-best results are highlighted in \highlightbox{best}{green} and \highlightbox{second}{blue}, respectively.
}
\label{tab:ablation}
\small
\resizebox{\textwidth}{!}{
\begin{tabular}{lcc cc cc cc}
\toprule
\multirow{2.5}{*}{Setting}
& \multicolumn{2}{c}{$\tau$-Bench-Return}
& \multicolumn{2}{c}{ALFWorld}
& \multicolumn{2}{c}{TextCraft}
& \multirow{2.5}{*}{Avg. Score $\uparrow$}
& \multirow{2.5}{*}{$\Delta$} \\
\cmidrule(lr){2-3}
\cmidrule(lr){4-5}
\cmidrule(lr){6-7}
& Score $\uparrow$ & Eff. $\uparrow$
& Score $\uparrow$ & Eff. $\uparrow$
& Score $\uparrow$ & Eff. $\uparrow$
& & \\
\midrule

ADIAS (Full)
& \cellcolor{best}81.3
& \cellcolor{best}6.35
& \cellcolor{best}94.0
& \cellcolor{best}8.95
& \cellcolor{best}91.0
& \cellcolor{best}9.01
& \cellcolor{best}88.8
& -- \\

\addlinespace[2pt]
\multicolumn{9}{l}{\textit{Ablating evidence for issue-state construction}} \\

\quad w/o External Prior
& 75.0
& 5.64
& \cellcolor{second}82.8
& \cellcolor{second}5.27
& 61.0
& 3.10
& 72.9
& -15.9 (17.9\%) \\

\quad w/o Round-Level Diagnosis
& \cellcolor{second}78.1
& \cellcolor{second}5.66
& 63.4
& 3.41
& \cellcolor{second}76.0
& \cellcolor{second}5.80
& 72.5
& -16.3 (18.4\%)\\

\addlinespace[2pt] 
\multicolumn{9}{l}{\textit{Replacing persistent issue state with a raw archive}} \\ 
\quad w/ Archive-Wide Synthesis
& 65.6
& 4.43
& 60.4
& 3.25
& 32.0
& 1.27
& 52.7
& -36.1 (40.7\%)\\

\addlinespace[2pt]
\multicolumn{9}{l}{\textit{Decoupling issue state from optimization control}} \\

\quad w/ Best-Candidate Revision
& 71.9
& 5.10
& 75.4
& 4.54
& 74.0
& 5.03
& \cellcolor{second}73.8
& -15.0 (16.9\%)\\

\quad w/ Latest-Candidate Continuation
& 62.5
& 4.37
& 41.0
& 1.81
& 60.0
& 3.53
& 54.5
& -34.3 (38.6\%)\\

\bottomrule
\end{tabular}
}
\end{table*}

\subsection{Qualitative Analysis}
\label{sec:qualitative}

\begin{figure*}[t]
    \centering
    \includegraphics[width=.85\textwidth]{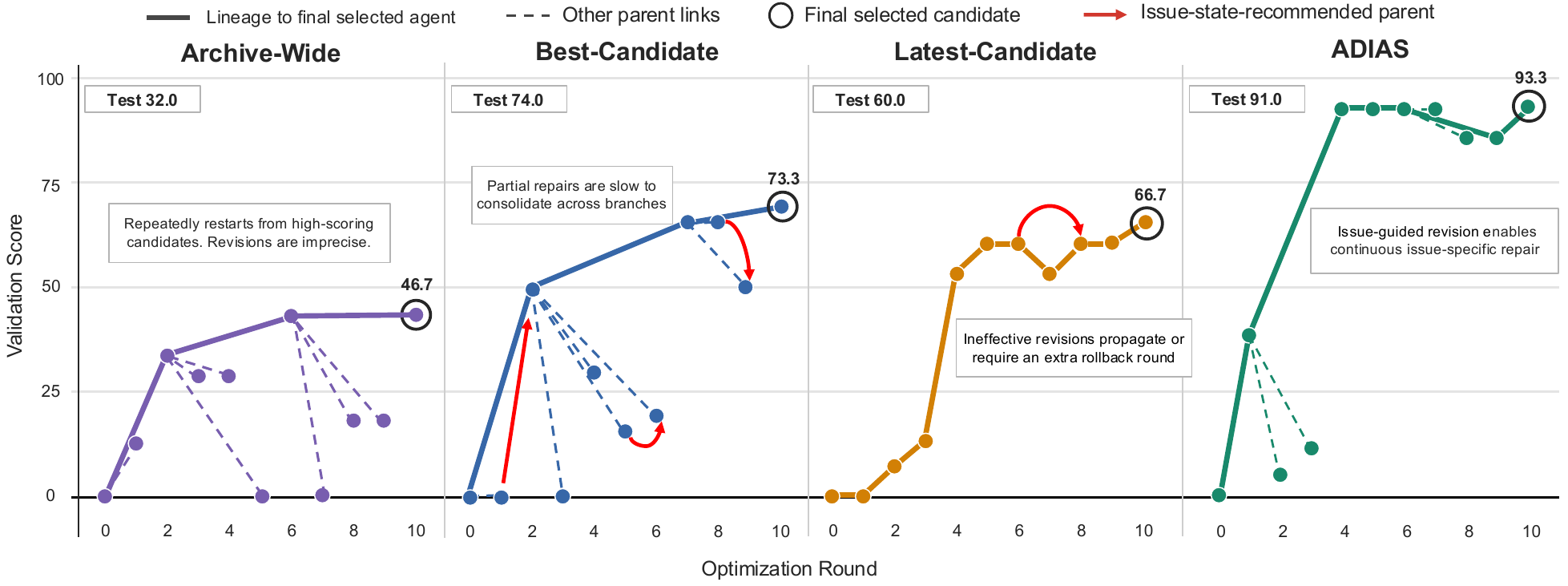}
    \caption{
    Comparison of optimization processes on the TextCraft benchmark. 
    }
    \label{fig:textcraft-lineage}
\end{figure*}

We inspect optimization traces to understand how issue state shapes the harness evolution process. The behaviors described below recur across all three benchmarks in Table \ref{tab:ablation}. Figure~\ref{fig:textcraft-lineage} illustrates a representative example of the optimization process on the TextCraft benchmark.

\textbf{A raw candidate archive provides weaker control signals than persistent issue state in long-horizon, sparse-feedback environments.}
We observe an anchoring mechanism similar to that described in \citet{lee2026meta}: optimization often starts from the highest-scoring candidate or the latest accepted one, and then attempts to summarize a new improvement direction from the surrounding archive. This strategy is less reliable when useful evidence is distributed across long interaction trajectories and sparse outcome signals. For example, in TextCraft (Figure \ref{fig:textcraft-lineage}), later rounds repeatedly return to previous high-score generations 2 and 6 and explore prompt restructuring, admissible-command matching, and command-selection heuristics, but the underlying missing-quantity error is never localized (get quartz vs. get 1 quartz). Reconstructing issues from candidate records each round therefore obscures effective repairs, whereas persistent issue state preserves explicit issue--intervention--outcome histories to guide revision.

\textbf{Best-Candidate Revision makes partial repairs slower to consolidate across branches.}
When a revision partially addresses an issue but lowers the aggregate score, its code-level changes are not inherited by the next candidate. The optimizer has to recover the useful insight from history and attempt a related repair again on a higher-scoring parent. In TextCraft, early branches introduce repairs of command recovery and crafting guidance, but their validation scores remain below generation 2, causing subsequent rounds to repeatedly return to that candidate and abandon parts of the previous implementations. Only after several alternative revisions, generation 7 combines effective revisions and improves validation from 50.0 to 66.7. The pattern is stronger in ALFWorld, where generations 4-10 repeatedly branch from generation 3 while separately targeting object identity, wrong-object recovery, action validation, and planner behavior. Although these interventions remain in history, their useful components must be repeatedly attributed and reimplemented across branches, making consolidation slower and less direct than with issue-specific intervention outcomes.

\textbf{Latest-Candidate Continuation makes ineffective revisions more likely to propagate, either accumulating errors or consuming additional rounds to recover.}
For example, in TextCraft, generation 7 introduces fallback exploration and reduces validation success from 60.0 to 53.3. Although the issue state recommends returning to generation 6, the fixed policy makes generation 8 inherit generation 7. The next round removes the ineffective modification merely to recover the previous validation score, which spends one additional round on rollback rather than further improvement. In contrast, ADIAS would return to generation 6 and resume optimization from a more suitable parent. The same problem is more pronounced in ALFWorld. After generation 8 is identified as unsuccessful, the issue state recommends generation 4 as the next parent, yet generation 9 is still derived from generation 8. In this case, the ineffective revision remains instead of being bypassed, which makes subsequent candidates more likely to inherit its errors. 

\textbf{ADIAS enables continuous issue-specific repair by selecting a more appropriate starting point and revision direction in each round.}
In TextCraft, ADIAS identifies the recurring command failure and repairs it by normalization. The intervention and its outcome are then written back to the persistent issue state, which allows subsequent rounds to retain the successful repair and continue addressing the remaining grounding and recovery issues. When modifications at generation 9 fail to improve task success rate but contribute to other issues, such as interaction efficiency, ADIAS still treats round 9 as a more suitable parent to accumulate partial revision. A similar process appears in ALFWorld, where optimization progresses from broad planning failures to more specific issues in action correction, planning heuristics, and target-object extraction. Across all benchmarks, ADIAS uses the active issue, previous interventions, and their observed outcomes to determine both where to restart and what to modify, which supports a continuous repair process.

\section{Conclusion}

We introduced issue-centric agent optimization, which reframes agent harness self-evolution as a cumulative optimization process rather than repeated candidate generation. We instantiated this formulation in ADIAS, a framework for automated full-code harness evolution that combines persistent issue state maintenance and issue-guided optimization. Across five interactive benchmarks, ADIAS outperforms the strongest baseline by 25.2\% on average and generalizes consistently across four backbone models. Controlled ablations and qualitative analyses further show that both persistent issue-level organization and its role as an optimization control state are necessary. These results suggest that future self-evolving agentic systems should treat optimization progress as an explicit, persistent, and operational state.






\bibliography{iclr2027_conference}

\begin{thebibliography}{53}
\providecommand{\natexlab}[1]{#1}
\providecommand{\url}[1]{\texttt{#1}}
\expandafter\ifx\csname urlstyle\endcsname\relax
  \providecommand{\doi}[1]{doi: #1}\else
  \providecommand{\doi}{doi: \begingroup \urlstyle{rm}\Url}\fi

\bibitem[Agrawal et~al.(2026)Agrawal, Tan, Soylu, Ziems, Khare, Opsahl-Ong, Singhvi, Shandilya, Ryan, Jiang, Potts, Sen, Dimakis, Stoica, Klein, Zaharia, and Khattab]{agrawal2026gepa}
Lakshya~A Agrawal, Shangyin Tan, Dilara Soylu, Noah Ziems, Rishi Khare, Krista Opsahl-Ong, Arnav Singhvi, Herumb Shandilya, Michael~J Ryan, Meng Jiang, Christopher Potts, Koushik Sen, Alex Dimakis, Ion Stoica, Dan Klein, Matei Zaharia, and Omar Khattab.
\newblock {GEPA}: Reflective prompt evolution can outperform reinforcement learning.
\newblock In \emph{The Fourteenth International Conference on Learning Representations}, 2026.
\newblock URL \url{https://openreview.net/forum?id=RQm2KQTM5r}.

\bibitem[ang Gao et~al.(2026)ang Gao, Geng, Hua, Hu, Juan, Liu, Liu, Qiu, Qi, Ren, Wu, WANG, Xiao, Zhou, Zhang, Zhang, Xiang, Fang, Zhao, Liu, Qian, Wang, Hu, Wang, Wu, Ji, and Wang]{gaosurvey2026}
Huan ang Gao, Jiayi Geng, Wenyue Hua, Mengkang Hu, Xinzhe Juan, Hongzhang Liu, Shilong Liu, Jiahao Qiu, Xuan Qi, Qihan Ren, Yiran Wu, Hongru WANG, Han Xiao, Yuhang Zhou, Shaokun Zhang, Jiayi Zhang, Jinyu Xiang, Yixiong Fang, Qiwen Zhao, Dongrui Liu, Cheng Qian, Zhenhailong Wang, Minda Hu, Huazheng Wang, Qingyun Wu, Heng Ji, and Mengdi Wang.
\newblock A survey of self-evolving agents: What, when, how, and where to evolve on the path to artificial super intelligence.
\newblock \emph{Transactions on Machine Learning Research}, 2026.
\newblock ISSN 2835-8856.
\newblock URL \url{https://openreview.net/forum?id=CTr3bovS5F}.
\newblock Survey Certification.

\bibitem[Cai et~al.(2026)Cai, Zhang, Jia, Zheng, Xue, Song, Tian, and Guo]{cai2026moss}
Qianshu Cai, Yonggang Zhang, Xianzhang Jia, Huajiang Zheng, Wei Xue, Jun Song, Xinmei Tian, and Yike Guo.
\newblock Moss: Self-evolution through source-level rewriting in autonomous agent systems.
\newblock \emph{arXiv preprint arXiv:2605.22794}, 2026.

\bibitem[Chen et~al.(2026)Chen, Lu, Zhao, Meng, Teng, Li, Li, Liu, Liang, Zhang, et~al.]{chen2026harnessx}
Tingyang Chen, Shuo Lu, Kang Zhao, Weicheng Meng, Hanlin Teng, Tianhao Li, Chao Li, Xule Liu, Jian Liang, Zhizhong Zhang, et~al.
\newblock Harnessx: A composable, adaptive, and evolvable agent harness foundry.
\newblock \emph{arXiv preprint arXiv:2606.14249}, 2026.

\bibitem[{DeepSeek}(2026)]{deepseek2026v4}
{DeepSeek}.
\newblock {DeepSeek V4 Preview Release}.
\newblock \url{https://api-docs.deepseek.com/news/news260424}, 2026.
\newblock Accessed: 2026-07.

\bibitem[Fernando et~al.(2024)Fernando, Banarse, Michalewski, Osindero, and Rockt{\"a}schel]{fernandopromptbreeder}
Chrisantha Fernando, Dylan~Sunil Banarse, Henryk Michalewski, Simon Osindero, and Tim Rockt{\"a}schel.
\newblock Promptbreeder: Self-referential self-improvement via prompt evolution.
\newblock In \emph{Forty-first International Conference on Machine Learning}, 2024.
\newblock URL \url{https://openreview.net/forum?id=9ZxnPZGmPU}.

\bibitem[Guo et~al.(2024)Guo, Wang, Guo, Li, Song, Tan, Liu, Bian, and Yang]{guo2024connecting}
Qingyan Guo, Rui Wang, Junliang Guo, Bei Li, Kaitao Song, Xu~Tan, Guoqing Liu, Jiang Bian, and Yujiu Yang.
\newblock Connecting large language models with evolutionary algorithms yields powerful prompt optimizers.
\newblock In \emph{The Twelfth International Conference on Learning Representations}, 2024.
\newblock URL \url{https://openreview.net/forum?id=ZG3RaNIsO8}.

\bibitem[Hu et~al.(2025)Hu, Lu, and Clune]{hu2025automated}
Shengran Hu, Cong Lu, and Jeff Clune.
\newblock Automated design of agentic systems.
\newblock In \emph{The Thirteenth International Conference on Learning Representations}, 2025.
\newblock URL \url{https://openreview.net/forum?id=t9U3LW7JVX}.

\bibitem[Huang et~al.(2024)Huang, Liu, Chen, Wang, Wang, Lian, Wang, Tang, and Chen]{huang2024understanding}
Xu~Huang, Weiwen Liu, Xiaolong Chen, Xingmei Wang, Hao Wang, Defu Lian, Yasheng Wang, Ruiming Tang, and Enhong Chen.
\newblock Understanding the planning of llm agents: A survey.
\newblock \emph{arXiv preprint arXiv:2402.02716}, 2024.

\bibitem[Huang et~al.(2026)Huang, Wang, Bao, Ma, Luo, Nian, Zhuang, Liu, Zhao, and Zhang]{huang2026memoharness}
Yue Huang, Wenjie Wang, Han Bao, Yuchen Ma, Xiaonan Luo, Yi~Nian, Haomin Zhuang, Zheyuan Liu, Yue Zhao, and Xiangliang Zhang.
\newblock Memoharness: Agent harnesses that learn from experience.
\newblock \emph{arXiv preprint arXiv:2607.14159}, 2026.

\bibitem[Khattab et~al.(2024)Khattab, Singhvi, Maheshwari, Zhang, Santhanam, A, Haq, Sharma, Joshi, Moazam, Miller, Zaharia, and Potts]{khattab2024dspy}
Omar Khattab, Arnav Singhvi, Paridhi Maheshwari, Zhiyuan Zhang, Keshav Santhanam, Sri~Vardhamanan A, Saiful Haq, Ashutosh Sharma, Thomas Joshi, Hanna Moazam, Heather Miller, Matei Zaharia, and Christopher Potts.
\newblock Dspy: Compiling declarative language model calls into state-of-the-art pipelines.
\newblock In B.~Kim, Y.~Yue, S.~Chaudhuri, K.~Fragkiadaki, M.~Khan, and Y.~Sun (eds.), \emph{International Conference on Learning Representations}, volume 2024, pp.\  54928--54958, 2024.
\newblock URL \url{https://proceedings.iclr.cc/paper_files/paper/2024/file/f1cf02ce09757f57c3b93c0db83181e0-Paper-Conference.pdf}.

\bibitem[Lee et~al.(2026)Lee, Nair, Zhang, Lee, Khattab, and Finn]{lee2026meta}
Yoonho Lee, Roshen Nair, Qizheng Zhang, Kangwook Lee, Omar Khattab, and Chelsea Finn.
\newblock Meta-harness: End-to-end optimization of model harnesses.
\newblock \emph{arXiv preprint arXiv:2603.28052}, 2026.

\bibitem[Li et~al.(2026)Li, Li, Wu, Liao, Hao, Shao, and Xu]{li2026agentswift}
Yu~Li, Lehui Li, Zhihao Wu, Qingmin Liao, Jianye Hao, Kun Shao, and Fengli Xu.
\newblock Agentswift: Efficient llm agent design via value-guided hierarchical search.
\newblock In \emph{Proceedings of the AAAI Conference on Artificial Intelligence}, volume~40, pp.\  31843--31851, 2026.

\bibitem[Li et~al.(2024)Li, Xu, Mei, Hua, Rama, Raheja, Wang, Zhu, and Zhang]{li2024autoflow}
Zelong Li, Shuyuan Xu, Kai Mei, Wenyue Hua, Balaji Rama, Om~Raheja, Hao Wang, He~Zhu, and Yongfeng Zhang.
\newblock Autoflow: Automated workflow generation for large language model agents.
\newblock \emph{arXiv preprint arXiv:2407.12821}, 2024.

\bibitem[Lin et~al.(2026{\natexlab{a}})Lin, Liu, Pan, Lin, Dou, Xi, Huang, Yan, Han, Gui, et~al.]{lin2026agentic}
Jiahang Lin, Shichun Liu, Chengjun Pan, Lizhi Lin, Shihan Dou, Zhiheng Xi, Xuanjing Huang, Hang Yan, Zhenhua Han, Tao Gui, et~al.
\newblock Agentic harness engineering: Observability-driven automatic evolution of coding-agent harnesses.
\newblock \emph{arXiv preprint arXiv:2604.25850}, 2026{\natexlab{a}}.

\bibitem[Lin et~al.(2026{\natexlab{b}})Lin, Lu, Shi, He, Mao, Zhang, Wu, Tang, Liu, Dai, et~al.]{lin2026position}
Minhua Lin, Hanqing Lu, Zhan Shi, Bing He, Rui Mao, Zhiwei Zhang, Zongyu Wu, Xianfeng Tang, Hui Liu, Zhenwei Dai, et~al.
\newblock Position: Agentic evolution is the path to evolving llms.
\newblock \emph{arXiv preprint arXiv:2602.00359}, 2026{\natexlab{b}}.

\bibitem[Liu et~al.(2024)Liu, Yu, Zhang, Xu, Lei, Lai, Gu, Ding, Men, Yang, et~al.]{liu2024agentbench}
Xiao Liu, Hao Yu, Hanchen Zhang, Yifan Xu, Xuanyu Lei, Hanyu Lai, Yu~Gu, Hangliang Ding, Kaiwen Men, Kejuan Yang, et~al.
\newblock Agentbench: Evaluating llms as agents.
\newblock In \emph{International Conference on Learning Representations}, volume 2024, pp.\  52989--53046, 2024.

\bibitem[Lou et~al.(2026)Lou, L{\'a}zaro-Gredilla, Dedieu, Wendelken, Lehrach, and Murphy]{lou2026autoharness}
Xinghua Lou, Miguel L{\'a}zaro-Gredilla, Antoine Dedieu, Carter Wendelken, Wolfgang Lehrach, and Kevin~P Murphy.
\newblock Autoharness: improving llm agents by automatically synthesizing a code harness.
\newblock \emph{arXiv preprint arXiv:2603.03329}, 2026.

\bibitem[Madžar \& Mekterović(2026)Madžar and Mekterović]{maksim2026}
Maksim Madžar and Igor Mekterović.
\newblock Automated design of agentic systems: A survey of algorithms for searching, optimizing, and evolving llm agents, workflows, and prompts.
\newblock \emph{Preprints}, June 2026.
\newblock \doi{10.20944/preprints202606.0238.v1}.
\newblock URL \url{https://doi.org/10.20944/preprints202606.0238.v1}.

\bibitem[Nie et~al.(2026)Nie, Zhang, Song, Cai, Yu, Guo, Tian, and Han]{nie2026tthe}
Jun Nie, Yonggang Zhang, Jun Song, Qianshu Cai, Dahai Yu, Yike Guo, Xinmei Tian, and Bo~Han.
\newblock Tthe: Test-time harness evolution.
\newblock \emph{arXiv preprint arXiv:2607.08124}, 2026.

\bibitem[Ning et~al.(2026)Ning, Tieu, Fu, Wei, Li, Bei, Zou, Ai, Liu, Li, et~al.]{ning2026code}
Xuying Ning, Katherine Tieu, Dongqi Fu, Tianxin Wei, Zihao Li, Yuanchen Bei, Jiaru Zou, Mengting Ai, Zhining Liu, Ting-Wei Li, et~al.
\newblock Code as agent harness.
\newblock \emph{arXiv preprint arXiv:2605.18747}, 2026.

\bibitem[{OpenAI}(2026)]{openai2026gpt54}
{OpenAI}.
\newblock {GPT-5.4 Model}.
\newblock \url{https://developers.openai.com/api/docs/models/gpt-5.4}, 2026.
\newblock Accessed: 2026-07.

\bibitem[Prasad et~al.(2024)Prasad, Koller, Hartmann, Clark, Sabharwal, Bansal, and Khot]{prasad2024adapt}
Archiki Prasad, Alexander Koller, Mareike Hartmann, Peter Clark, Ashish Sabharwal, Mohit Bansal, and Tushar Khot.
\newblock Adapt: As-needed decomposition and planning with language models.
\newblock In \emph{Findings of the Association for Computational Linguistics: NAACL 2024}, pp.\  4226--4252, 2024.

\bibitem[Pryzant et~al.(2023)Pryzant, Iter, Li, Lee, Zhu, and Zeng]{pryzant2023automatic}
Reid Pryzant, Dan Iter, Jerry Li, Yin Lee, Chenguang Zhu, and Michael Zeng.
\newblock Automatic prompt optimization with “gradient descent” and beam search.
\newblock In \emph{Proceedings of the 2023 conference on empirical methods in natural language processing}, pp.\  7957--7968, 2023.

\bibitem[Saad-Falcon et~al.(2024)Saad-Falcon, Lafuente, Natarajan, Maru, Todorov, Guha, Buchanan, Chen, Guha, R{\'e}, et~al.]{saad2024archon}
Jon Saad-Falcon, Adrian~Gamarra Lafuente, Shlok Natarajan, Nahum Maru, Hristo Todorov, Etash Guha, E~Kelly Buchanan, Mayee Chen, Neel Guha, Christopher R{\'e}, et~al.
\newblock Archon: An architecture search framework for inference-time techniques.
\newblock \emph{arXiv preprint arXiv:2409.15254}, 2024.

\bibitem[Shang et~al.(2025)Shang, Li, Zhao, Ma, Liu, Xu, and Li]{shang2025agentsquare}
Yu~Shang, Yu~Li, Keyu Zhao, Likai Ma, Jiahe Liu, Fengli Xu, and Yong Li.
\newblock Agentsquare: Automatic llm agent search in modular design space.
\newblock In \emph{International Conference on Learning Representations}, volume 2025, pp.\  3841--3865, 2025.

\bibitem[Shinn et~al.(2023)Shinn, Cassano, Gopinath, Narasimhan, and Yao]{shinn2023reflexion}
Noah Shinn, Federico Cassano, Ashwin Gopinath, Karthik Narasimhan, and Shunyu Yao.
\newblock Reflexion: Language agents with verbal reinforcement learning.
\newblock \emph{Advances in neural information processing systems}, 36:\penalty0 8634--8652, 2023.

\bibitem[Shridhar et~al.(2021)Shridhar, Yuan, Cote, Bisk, Trischler, and Hausknecht]{shridharalfworld}
Mohit Shridhar, Xingdi Yuan, Marc-Alexandre Cote, Yonatan Bisk, Adam Trischler, and Matthew Hausknecht.
\newblock Alfworld: Aligning text and embodied environments for interactive learning.
\newblock In \emph{International Conference on Learning Representations}, 2021.

\bibitem[{Tencent Hy}(2026)]{tencent2026hy3}
{Tencent Hy}.
\newblock {Introducing Hy3}.
\newblock \url{https://hy.tencent.com/research/hy3}, 2026.
\newblock Accessed: 2026-07.

\bibitem[Wang et~al.(2024{\natexlab{a}})Wang, Ma, Feng, Zhang, Yang, Zhang, Chen, Tang, Chen, Lin, et~al.]{wang2024survey}
Lei Wang, Chen Ma, Xueyang Feng, Zeyu Zhang, Hao Yang, Jingsen Zhang, Zhiyuan Chen, Jiakai Tang, Xu~Chen, Yankai Lin, et~al.
\newblock A survey on large language model based autonomous agents.
\newblock \emph{Frontiers of Computer Science}, 18\penalty0 (6):\penalty0 186345, 2024{\natexlab{a}}.

\bibitem[Wang et~al.(2026{\natexlab{a}})Wang, Shi, Li, Li, Yu, Yang, Panaganti, Mi, Zhou, et~al.]{wang2026harness}
Ruhan Wang, Yucheng Shi, Zongxia Li, Zhongzhi Li, Yue Yu, Junyao Yang, Kishan Panaganti, Haitao Mi, Dongruo Zhou, et~al.
\newblock Harness handbook: Making evolving agent harnesses readable, navigable, and editable.
\newblock \emph{arXiv preprint arXiv:2607.13285}, 2026{\natexlab{a}}.

\bibitem[Wang et~al.(2022)Wang, Jansen, C{\^o}t{\'e}, and Ammanabrolu]{wang2022scienceworld}
Ruoyao Wang, Peter Jansen, Marc-Alexandre C{\^o}t{\'e}, and Prithviraj Ammanabrolu.
\newblock Scienceworld: Is your agent smarter than a 5th grader?
\newblock In \emph{Proceedings of the 2022 Conference on Empirical Methods in Natural Language Processing}, pp.\  11279--11298, 2022.

\bibitem[Wang et~al.(2026{\natexlab{b}})Wang, Pi{\k{e}}kos, Nanbo, Laakom, Chen, Ostaszewski, Zhuge, and Schmidhuber]{wang2025huxley}
Wenyi Wang, Piotr Pi{\k{e}}kos, Li~Nanbo, Firas Laakom, Yimeng Chen, Mateusz Ostaszewski, Mingchen Zhuge, and J{\"u}rgen Schmidhuber.
\newblock Huxley-g{\"odel} machine: Human-level coding agent development by an approximation of the optimal self-improving machine.
\newblock In \emph{The Fourteenth International Conference on Learning Representations}, 2026{\natexlab{b}}.
\newblock URL \url{https://openreview.net/forum?id=T0EiEuhOOL}.

\bibitem[Wang et~al.(2024{\natexlab{b}})Wang, Li, Wang, Bai, Luo, Zhang, Jojic, Xing, and Hu]{wang2024promptagent}
Xinyuan Wang, Chenxi Li, Zhen Wang, Fan Bai, Haotian Luo, Jiayou Zhang, Nebojsa Jojic, Eric Xing, and Zhiting Hu.
\newblock Promptagent: Strategic planning with language models enables expert-level prompt optimization.
\newblock In \emph{International Conference on Learning Representations}, volume 2024, pp.\  23967--24001, 2024{\natexlab{b}}.

\bibitem[Wang et~al.(2026{\natexlab{c}})Wang, Zhu, Hu, Yuan, Chen, Senthil, Hajishirzi, Tsvetkov, Dasigi, and Xiao]{wang2026rethinking}
Yike Wang, Huaisheng Zhu, Zhengyu Hu, Yige Yuan, Zhengyu Chen, Shakti Senthil, Hannaneh Hajishirzi, Yulia Tsvetkov, Pradeep Dasigi, and Teng Xiao.
\newblock Rethinking the evaluation of harness evolution for agents.
\newblock \emph{arXiv preprint arXiv:2607.12227}, 2026{\natexlab{c}}.

\bibitem[Weng(2026)]{weng2026harness}
Lilian Weng.
\newblock Harness engineering for self-improvement.
\newblock \emph{lilianweng.github.io}, July 2026.
\newblock URL \url{https://lilianweng.github.io/posts/2026-07-04-harness/}.

\bibitem[Yang et~al.(2024)Yang, Wang, Lu, Liu, Le, Zhou, and Chen]{yang2024large}
Chengrun Yang, Xuezhi Wang, Yifeng Lu, Hanxiao Liu, Quoc~V Le, Denny Zhou, and Xinyun Chen.
\newblock Large language models as optimizers.
\newblock In \emph{International Conference on Learning Representations}, volume 2024, pp.\  12028--12068, 2024.

\bibitem[Yang et~al.(2026)Yang, Gong, Huang, Yang, Zhou, Huang, Li, Gao, Dai, Liu, et~al.]{yang2026skillopt}
Yifan Yang, Ziyang Gong, Weiquan Huang, Qihao Yang, Ziwei Zhou, Zisu Huang, Yan Li, Xuemei Gao, Qi~Dai, Bei Liu, et~al.
\newblock Skillopt: Executive strategy for self-evolving agent skills.
\newblock \emph{arXiv preprint arXiv:2605.23904}, 2026.

\bibitem[Yao et~al.(2022)Yao, Chen, Yang, and Narasimhan]{yao2022webshop}
Shunyu Yao, Howard Chen, John Yang, and Karthik Narasimhan.
\newblock Webshop: Towards scalable real-world web interaction with grounded language agents.
\newblock \emph{Advances in Neural Information Processing Systems}, 35:\penalty0 20744--20757, 2022.

\bibitem[Yao et~al.(2023)Yao, Zhao, Yu, Du, Shafran, Narasimhan, and Cao]{yao2023react}
Shunyu Yao, Jeffrey Zhao, Dian Yu, Nan Du, Izhak Shafran, Karthik Narasimhan, and Yuan Cao.
\newblock React: Synergizing reasoning and acting in language models.
\newblock In \emph{International Conference on Learning Representations (ICLR)}, 2023.

\bibitem[Yao et~al.(2025)Yao, Shinn, Razavi, and Narasimhan]{yao2025tau}
Shunyu Yao, Noah Shinn, Pedram Razavi, and Karthik~R Narasimhan.
\newblock $\tau$-bench: A benchmark for tool-agent-user interaction in real-world domains.
\newblock In \emph{The Thirteenth International Conference on Learning Representations}, 2025.

\bibitem[Yin et~al.(2025)Yin, Wang, Pan, Lin, Wan, and Wang]{yin2025godel}
Xunjian Yin, Xinyi Wang, Liangming Pan, Li~Lin, Xiaojun Wan, and William~Yang Wang.
\newblock G{\"o}del agent: A self-referential agent framework for recursively self-improvement.
\newblock In \emph{Proceedings of the 63rd Annual Meeting of the Association for Computational Linguistics (Volume 1: Long Papers)}, pp.\  27890--27913, 2025.

\bibitem[Yue et~al.(2026)Yue, Bhandari, Ko, Patel, Lin, Zhou, Gao, Chen, and Pan]{yue2026static}
Ling Yue, Kushal~Raj Bhandari, Ching-Yun Ko, Dhaval Patel, Shuxin Lin, Nianjun Zhou, Jianxi Gao, Pin-Yu Chen, and Shaowu Pan.
\newblock From static templates to dynamic runtime graphs: a survey of workflow optimization for llm agents.
\newblock \emph{arXiv preprint arXiv:2603.22386}, 2026.

\bibitem[{Z.ai}(2026)]{zai2026glm52}
{Z.ai}.
\newblock {GLM-5.2 Model}.
\newblock \url{https://docs.bigmodel.cn/cn/guide/models/text/glm-5.2}, 2026.
\newblock Accessed: 2026-07.

\bibitem[Zhang et~al.(2025{\natexlab{a}})Zhang, Niu, Fang, Wang, Bai, and Wang]{zhang2025multi}
Guibin Zhang, Luyang Niu, Junfeng Fang, Kun Wang, Lei Bai, and Xiang Wang.
\newblock Multi-agent architecture search via agentic supernet.
\newblock In \emph{International Conference on Machine Learning}, pp.\  75834--75852. PMLR, 2025{\natexlab{a}}.

\bibitem[Zhang et~al.(2026{\natexlab{a}})Zhang, Hu, Lu, Lange, and Clune]{zhang2025darwin}
Jenny Zhang, Shengran Hu, Cong Lu, Robert Lange, and Jeff Clune.
\newblock Darwin g\"{o}del machine: Open-ended evolution of self-improving agents.
\newblock In C.~Vondrick, B.~Hariharan, C.~Raffel, L.~Pinto, D.~Yang, and A.~Faust (eds.), \emph{International Conference on Learning Representations}, volume 2026, pp.\  104223--104294, 2026{\natexlab{a}}.
\newblock URL \url{https://proceedings.iclr.cc/paper_files/paper/2026/file/aa5f5e6eb6f613ec412f1d948dfa21a5-Paper-Conference.pdf}.

\bibitem[Zhang et~al.(2026{\natexlab{b}})Zhang, Zhao, Yang, Foerster, Clune, Jiang, Devlin, and Shavrina]{zhang2026hyperagents}
Jenny Zhang, Bingchen Zhao, Wannan Yang, Jakob Foerster, Jeff Clune, Minqi Jiang, Sam Devlin, and Tatiana Shavrina.
\newblock Hyperagents.
\newblock \emph{arXiv preprint arXiv:2603.19461}, 2026{\natexlab{b}}.

\bibitem[Zhang et~al.(2025{\natexlab{b}})Zhang, Xiang, Yu, Teng, Chen, Chen, Zhuge, Cheng, Hong, Wang, et~al.]{zhang2025aflow}
Jiayi Zhang, Jinyu Xiang, Zhaoyang Yu, Fengwei Teng, Xionghui Chen, Jiaqi Chen, Mingchen Zhuge, Xin Cheng, Sirui Hong, Jinlin Wang, et~al.
\newblock Aflow: Automating agentic workflow generation.
\newblock In \emph{International Conference on Learning Representations}, volume 2025, pp.\  34040--34077, 2025{\natexlab{b}}.

\bibitem[Zhang et~al.(2026{\natexlab{c}})Zhang, Gu, Ruan, Song, Peng, Han, Xiang, Wang, Yang, Ouyang, et~al.]{zhang2026harnessing}
Jiayi Zhang, Yongfeng Gu, Jianhao Ruan, Maojia Song, Yiran Peng, Zhiguang Han, Jinyu Xiang, Zhitao Wang, Caiyin Yang, Yixi Ouyang, et~al.
\newblock Harnessing agentic evolution.
\newblock \emph{arXiv preprint arXiv:2605.13821}, 2026{\natexlab{c}}.

\bibitem[Zhang et~al.(2026{\natexlab{d}})Zhang, Hu, Upasani, Ma, Hong, Kamanuru, Rainton, Wu, Ji, Li, et~al.]{zhang2025agentic}
Qizheng Zhang, Changran Hu, Shubhangi Upasani, Boyuan Ma, Fenglu Hong, Vamsidhar Kamanuru, Jay Rainton, Chen Wu, Mengmeng Ji, Hanchen Li, et~al.
\newblock Agentic context engineering: Evolving contexts for self-improving language models.
\newblock In \emph{International Conference on Learning Representations}, 2026{\natexlab{d}}.

\bibitem[Zhao et~al.(2024)Zhao, Huang, Xu, Lin, Liu, and Huang]{zhao2024expel}
Andrew Zhao, Daniel Huang, Quentin Xu, Matthieu Lin, Yong-Jin Liu, and Gao Huang.
\newblock Expel: Llm agents are experiential learners.
\newblock In \emph{Proceedings of the AAAI Conference on Artificial Intelligence}, volume~38, pp.\  19632--19642, 2024.

\bibitem[Zhou et~al.(2023)Zhou, Muresanu, Han, Paster, Pitis, Chan, and Ba]{zhou2022large}
Yongchao Zhou, Andrei~Ioan Muresanu, Ziwen Han, Keiran Paster, Silviu Pitis, Harris Chan, and Jimmy Ba.
\newblock Large language models are human-level prompt engineers.
\newblock In \emph{The Eleventh International Conference on Learning Representations}, 2023.
\newblock URL \url{https://openreview.net/forum?id=92gvk82DE-}.

\bibitem[Zhuge et~al.(2024)Zhuge, Wang, Kirsch, Faccio, Khizbullin, and Schmidhuber]{zhuge2024gptswarm}
Mingchen Zhuge, Wenyi Wang, Louis Kirsch, Francesco Faccio, Dmitrii Khizbullin, and J{\"u}rgen Schmidhuber.
\newblock Gptswarm: Language agents as optimizable graphs.
\newblock In \emph{Forty-first International Conference on Machine Learning}, 2024.

\end{thebibliography}
\bibliographystyle{iclr2027_conference}

\appendix
\section{Interactive Agent}

Interactive environments make automated agent design substantially more difficult in three ways. 

First, interactive environments induce a larger and more task-dependent design space than single-input-output tasks.. Successful agents may require different combinations of observation processing, memory, planning, tool use, verification, and recovery mechanisms \citep{huang2024understanding,wang2024survey}. Full-code design provides the necessary expressiveness, but must search over a much larger and less structured space of executable implementations \citep{hu2025automated}. 

Second, interaction feedback is sparse, delayed, and difficult to attribute. Interactive agents may make many dependent decisions over multiple environment states before a final success or failure signal is observed \citep{liu2024agentbench,yao2025tau}. A scalar score reveals \emph{whether} an agent succeeded, but rarely explains \emph{why}: the failure may originate from an early misunderstanding, lost state, incorrect tool use, a policy violation, or ineffective recovery. Simply providing complete trajectories to an LLM is also problematic, as long and heavy traces contain substantial irrelevant context and make decisive failure evidence difficult to isolate. Effective design search therefore requires precise, evidence-grounded failure attribution.

Third, interactive evaluation is expensive, which makes search efficiency critical. Each candidate must be assessed through environment rollouts that may require extensive model calls and interaction steps \citep{yao2022webshop,yao2025tau}. Existing full-code design and self-improvement methods commonly retain and select evaluated agent variants based primarily on empirical task scores \citep{hu2025automated,zhang2025darwin,zhang2026hyperagents}. Without a persistent view of search progress, however, the designer may revisit unsuccessful modifications, overlook the recurrence of previously mitigated failures, and repeatedly spend evaluation budget on already explored directions. Efficient optimization therefore requires tracking what has been tried, what has improved or regressed, and which unresolved bottleneck should guide the next design round.

\begin{figure*}[t]
    \centering
    \includegraphics[width=\textwidth]{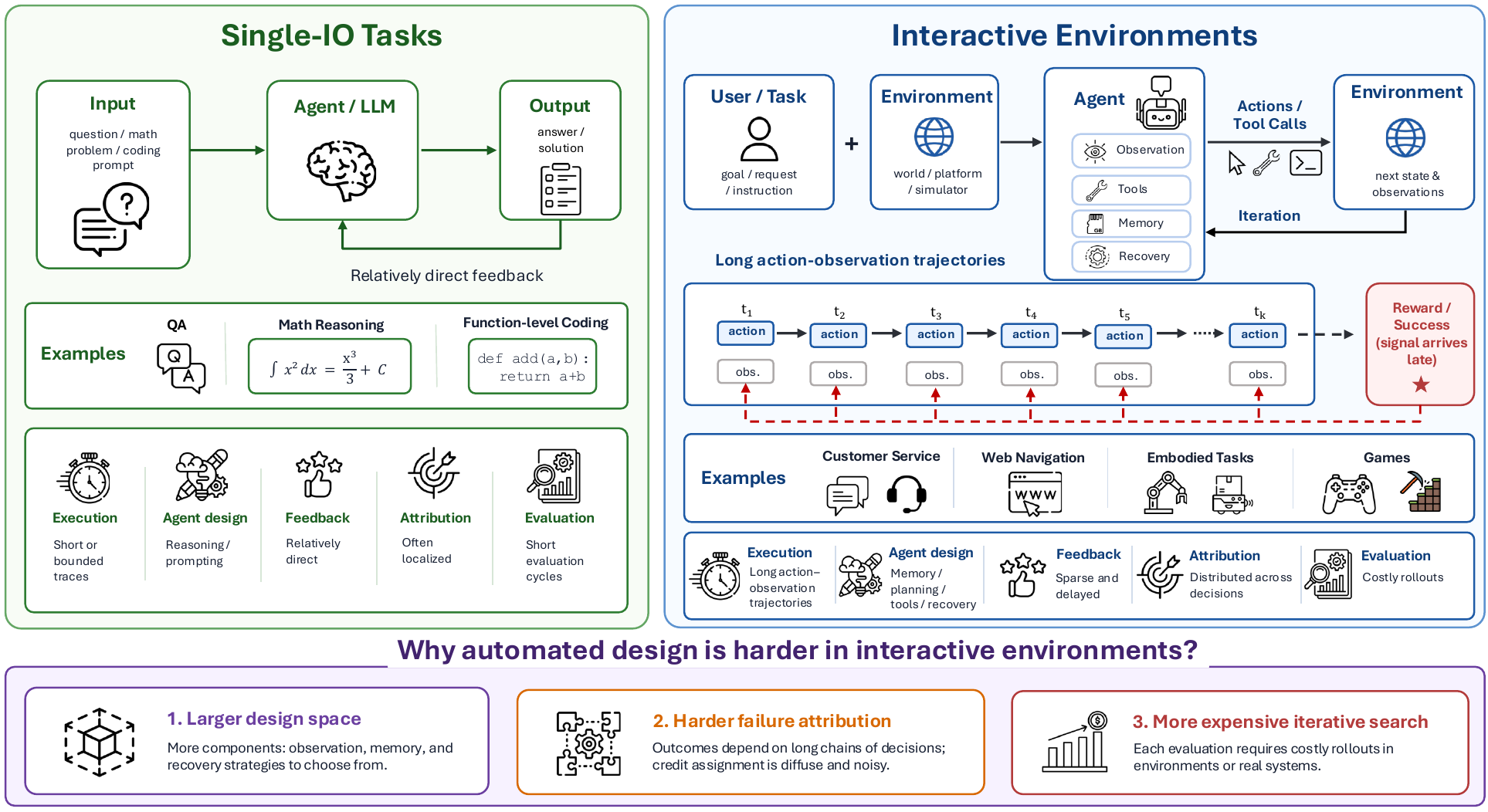}
    \caption{
    Task comparison between single-IO tasks and long-horizon interactive environments. 
    }
    \label{fig:task}
\end{figure*}

\section{Detailed Related Work}

Recent work has increasingly shifted from manually designing agentic systems toward automatically optimizing them from execution feedback. Automated Design of Agentic Systems (ADAS) formalizes this process as searching over agentic system designs under an evaluation objective \citep{maksim2026}. Related surveys study this broader direction through workflow optimization, self-evolving agents, and adaptive agent harnesses \citep{gaosurvey2026,yue2026static,ning2026code}. We organize prior work into three levels according to the expressiveness of the optimization target: prompt (text)-level, architecture-level, and full-code-level design.

\textbf{Prompt-level Design.}
Prompt-level methods optimize textual artifacts based on fixed models and an agent execution process. Early automatic prompt optimization methods search directly over natural-language instructions, using LLM-generated candidates \citep{zhou2022large}, scored optimization histories \citep{yang2024large}, textual gradients \citep{pryzant2023automatic}, evolutionary search \citep{guo2024connecting,fernandopromptbreeder}, or Monte Carlo tree search \citep{wang2024promptagent}. More recent methods exploit richer execution feedback: GEPA evolves prompts through trajectory reflection and Pareto-based selection, while DSPy optimizes instructions within modular language-model pipelines \citep{agrawal2026gepa,khattab2024dspy}.
ACE and SkillOpt further treat persistent context playbooks or skill documents as editable textual state, accumulating reusable procedures, heuristics, and failure-handling knowledge from experience \citep{zhang2025agentic,yang2026skillopt}.

\textbf{Architecture-level Design.}
Architecture-level methods design the agent's overall structure, while remaining constrained by a predefined graph, module set, or harness abstraction. Existing approaches broadly follow two directions. The first direction directly searches over agent architectures or workflows, including AutoFlow, AFlow, and GPTSwarm \citep{li2024autoflow,zhang2025aflow,zhuge2024gptswarm}.
Other methods compose predefined reasoning, planning, tool-use, memory, and inference-time components, as in AgentSquare, Archon, MaAS, and AgentSwift \citep{shang2025agentsquare,saad2024archon,zhang2025multi,li2026agentswift}.
More recent systems expose runtime harnesses themselves as structured, editable artifacts: A-Evolve maintains versioned knowledge, workflow, tool, and validation assets; AHE edits explicit harness components using trajectory observability; and HarnessX evolves compositions of typed harness primitives \citep{lin2026position,lin2026agentic,chen2026harnessx}.

\textbf{Full-code-level Design.}
At the most expressive level, agents are represented as executable programs and optimized directly in code space. \citet{hu2025automated} introduced this direction by using a fixed meta-agent to iteratively propose new task agent implementations and retain successful variants in an archive. DGM combines self-modification with open-ended evolutionary exploration over a branching archive \citep{zhang2025darwin}, while HGM further improves parent selection using estimated descendant potential \citep{wang2025huxley}.
In addition, Meta-Harness searches harness code using selective access to previous code, scores, and traces \citep{lee2026meta}; AutoHarness synthesizes executable harnesses from environment feedback \citep{lou2026autoharness}; MOSS performs source-level rewriting from curated failures \citep{cai2026moss}; and TTHE evolves harnesses at test time from unlabeled trajectories \citep{nie2026tthe}.
A further line of work makes parts of the improvement mechanism itself editable. G\"odel Agent supports self-referential modification of task behavior and improvement logic; DGM-H jointly evolves the task agent and meta-level modification mechanism; and AEvo revises the procedure or context that supports subsequent evolution \citep{yin2025godel,zhang2026hyperagents,zhang2026harnessing}.
These methods extend automated design from optimizing an agent artifact to optimizing how future agent modifications are designed.

\section{ADIAS Details}
\label{app:adiasdetails}

Algorithm~\ref{alg:adias} summarizes the complete procedure.

\subsection{External Prior Search}
\label{app:external-prior}

Full-code agent design exposes an extremely large space of possible implementations. However, the task protocol and public task-level knowledge often provide useful priors about likely interaction requirements and failure modes. ADIAS uses a one-shot external searcher to exploit this information before iterative design begins.

Given the task protocol $\mathcal{P}$, the external searcher retrieves public information such as benchmark descriptions, rule explanations, common failure modes, tool-use strategies, and candidate agent mechanisms. The resulting external prior $P$ is incorporated into the initial issue state as \emph{unverified task prior knowledge}. It is explicitly separated from empirical experience and is treated only as a set of design hypotheses. This distinction is important because retrieved information may be incomplete or unsuitable for the current agent. A prior is promoted into accumulated task experience only when supported by trajectory diagnosis, patch outcomes, runtime evidence, or score trends. The external searcher is executed exactly once and is not changed during late design iterations. We restrict retrieval to public task-level information and exclude held-out labels, instance-specific solutions, hidden evaluation data, and hard-coded benchmark outputs. Overall, external search narrows the effective cold-start search space without restricting the expressiveness of the underlying full-code design
space.

\begin{algorithm}[t]
\caption{Automated Design of Interactive Agentic Systems}
\label{alg:adias}
\begin{algorithmic}[1]
\REQUIRE Task protocol $\mathcal{P}$, initial agent $A_0$,
optimization round $T$

\STATE $P \leftarrow \mathrm{ExternalSearch}(\mathcal{P})$

\STATE $\mathcal{T}_0,M_0
\leftarrow \mathrm{Evaluate}(A_0)$
\STATE $E_0
\leftarrow \mathrm{InitializeIssueState}(P)$
\STATE $H_0
\leftarrow \mathrm{InitializeHistory}(A_0,\mathcal{T}_0,M_0)$

\FOR{$t=1,\ldots,T$}
    \STATE $D_t
    \leftarrow \mathrm{Diagnose}(\mathcal{T}_{t-1},M_{t-1})$
    \STATE $H_t
    \leftarrow \mathrm{UpdateHistory}({H}_{t-1}, A_{t-1},\mathcal{T}_{t-1},M_{t-1}, D_{t})$
    \STATE $E_t
    \leftarrow \mathrm{UpdateIssueState}(E_{t-1}, H_t)$
    \STATE $A_{p_t}, R_{t}
    \leftarrow \mathrm{Plan}(H_t, E_{t})$
    \STATE $A_t
    \leftarrow \mathrm{Modify}(A_{p_t},R_{t})$
    \STATE $\mathcal{T}_t,M_t
    \leftarrow \mathrm{Evaluate}(A_t)$
\ENDFOR

\STATE \textbf{return}
$\displaystyle
A^\star =
\arg\max_{A_i\in\mathcal{A}_T} {M}_{val}(A_i)$
\end{algorithmic}
\end{algorithm}

\subsection{Evidence-Grounded Failure Diagnosis}
\label{app:diagnosis}
A scalar evaluation score reveals whether an agent succeeds, but rarely identifies which design deficiency caused a failure. This attribution problem is difficult in interactive tasks because a single episode may contain many state transitions, tool calls, and dependent decisions. Directly placing all trajectories into the context of a designer model also introduces substantial irrelevant information.

ADIAS therefore introduces a dedicated diagnostic agent that analyzes all training trajectories collected in one optimization round. Rather than consuming complete trajectories as a flat prompt, the diagnostic agent interacts with a structured trajectory database through targeted query tools. It can inspect aggregate failure patterns, compare successful and failed episodes, retrieve compact action sequences or local step windows, search for repeated and invalid actions, and examine tool errors, visible final states, etc. The diagnostic agent uses these queries to iteratively examine failures based on trajectory evidence. The output is a diagnostic report
\begin{equation}
    D_t = (I_t, S_t, C_t),
\end{equation}
where $I_t$ denotes observed issue categories, $S_t$ contains supporting evidence, and $C_t$ describes the inferred causes. The objective is not merely to locate a failed action, but to identify recurring behavioral patterns that indicate deficiencies in the current agent design. Diagnosis is deliberately separated from modification, which means the diagnostic agent does not propose suggestions or implement code patches.

\subsection{Persistent Issue Control}
\label{subsec:issuecontrol}

This section describes how ADIAS operationalizes that state for full-code agent design. In our implementation, $E_t$ is maintained as a structured profile containing a global optimization summary and a set of issue records. Each issue record instantiates the fields in Section~\ref{sec:manager} with a concise issue description, priority, lifecycle status, evidence references, previously attempted interventions, observed outcomes, and the candidate generations in which the issue was observed or absent. The candidate archive remains available separately and stores complete programs, trajectories, scores, and parent relations. Thus, $E_t$ does not replace the archive; it provides an issue-indexed control layer over candidate-level history.

After each evaluation, the Issue Manager compares the diagnostic report with the existing issue records. Observations are associated with an existing issue when they describe the same underlying behavioral failure, affected capability, and execution context, even if their surface descriptions differ across trajectories. Otherwise, a new stable issue identity is created. Supporting evidence stores references to the relevant trajectory segments and diagnostic findings rather than only an aggregate score change. External task priors are handled as provisional hypotheses: they may initialize issue records, but do not become confirmed failures unless supported by observed behavior.

The persistent issue state controls which problem should be addressed and where optimization should continue, but it does not prescribe a fixed implementation. Conditioned on the repair plan, the designer may add, remove, or rewrite any agent-side logic, such as prompts, tools, memory, and task-specific modules. After the revised candidate is evaluated, its diagnostic evidence and measured outcomes are written back to the corresponding issue records, forming an issue--intervention--outcome chain for the next round. This separation preserves the expressiveness of full-code optimization while supporting continuous repair across candidate generations.

\section{Experimental Details}
\label{app:experimentaldetails}

\begin{table*}[t]
\centering
\caption{
Summary of the five interactive benchmarks used in our experiments.
}
\label{tab:benchmark_summary}
\small
\renewcommand{\arraystretch}{1.12}

\resizebox{\textwidth}{!}{
\begin{tabular}{lllccc}
\toprule
Benchmark
& Domain
& \multicolumn{1}{c}{Capability}
& \#Train
& \#Val
& \#Test \\
\midrule

$\tau$-Bench-Return
& Customer Service
& Policy-aware tool use and user interaction
& 55 & 15 & 32 \\

$\tau$-Bench Retail
& Customer Service
& Policy-aware tool use and user interaction
& 500 & 20 & 115 \\

ALFWorld
& Embodied AI
& Long-horizon planning and state tracking
& 153 & 15 & 134 \\

TextCraft
& Game
& Hierarchical decomposition and dependency planning
& 60 & 15 & 100 \\

WebShop
& Web
& Information seeking and constraint-aware navigation
& 84 & 25 & 50 \\

ScienceWorld
& Scientific Scenes
& Grounded procedural and scientific reasoning
& 66 & 22 & 102 \\

\bottomrule
\end{tabular}
}
\end{table*}

\subsection{Benchmark Details}
\label{app:benchmarks}

To control computational costs, we construct representative task subsets for the main experiments. We preserve the native task structure and evaluator of each benchmark, and select subsets according to task semantics or trajectory complexity rather than sampling arbitrary individual episodes. 

\textbf{$\tau$-Bench.}
$\tau$-Bench evaluates agents in tool--agent--user interactions, requiring them to communicate with a simulated user, comply with domain-specific policies, and invoke tools that modify an underlying database state \citep{yao2025tau}. We use the retail domain at two evaluation scales.

For the main experiments, we use all qualified return-related tasks, comprising 55 training, 15 validation, and 32 test examples. Return requests form a coherent but nontrivial intent family: agents must identify the relevant order and items, verify policy constraints, collect missing user information, and execute the appropriate tool sequence. We adopt this setting primarily for computational efficiency. Restricting the optimization set to one complete intent family enables controlled comparison under a feasible and uniform budget.

To verify that the conclusions are not specific to return requests, we additionally repeat the comparison on the full $\tau$-Bench retail distribution, containing 500 training, 20 validation, and 115 test tasks. This expanded setting covers heterogeneous customer-service requests beyond returns and substantially increases the cost and diversity of optimization. The purpose is to show that the return subset provides a cost-efficient proxy for controlled comparison rather than an artificially favorable evaluation setting.

\textbf{ALFWorld.}
ALFWorld uses text-based environments to simulate embodied household tasks and evaluates instruction understanding, planning, navigation, and object interaction \citep{shridharalfworld}. For example, the \emph{Clean-and-Place} task family requires an agent to locate a target object, identify and navigate to the required locations, clean the object, and place it in a specified place. Successful execution therefore requires long-horizon planning, environment state tracking, and recovery from incorrect intermediate actions. We randomly sample 153 training and 15 validation examples from the \emph{Clean-and-Place} task family, together with 30 held-out \emph{Clean-and-Place} examples for in-distribution testing. To evaluate whether the optimized agent generalizes beyond the task family used during optimization, we additionally evaluate it on the official ALFWorld test set containing 134 examples across all task families. The results are based on the official test set by default. 

\textbf{TextCraft.}
TextCraft is a text-based gaming environment designed to evaluate compositional planning over Minecraft-inspired crafting recipes \citep{prasad2024adapt}. Given a target item and a set of available crafting operations, the agent must reason recursively about prerequisite materials and execute a valid sequence of gathering and crafting actions. Recipe depth naturally controls task complexity, with deeper recipes requiring longer chains of dependent subgoals. To obtain a complexity-balanced subset, we stratify examples by recipe depth and sample tasks from depths 2, 3, and 4. Our final split contains 60 training and 15 validation examples, where examples are evenly distributed across the three recipe depths. We use the official test set with 100 samples. 

\textbf{WebShop.}
WebShop is a simulated e-commerce environment in which an agent must satisfy a natural-language shopping instruction by searching products, navigating result and product pages, inspecting attributes, and selecting an appropriate item for purchase \citep{yao2022webshop}. The environment evaluates information seeking, constraint tracking, long-horizon web navigation, and comparison among multiple candidate products. To emphasize genuinely interactive tasks, we construct a subset from successful human trajectories and retain instances requiring more than 10 non-redundant interaction steps. This filtering removes short or trivial purchases and focuses evaluation on tasks involving sustained navigation and decision-making. The resulting subset contains 84 training, 25 validation, and 50 test instances.

\textbf{ScienceWorld.}
ScienceWorld evaluates scientific reasoning in an interactive text environment grounded in elementary-school science concepts \citep{wang2022scienceworld}. Agents must understand a scientific objective and carry out a valid sequence of observations, navigation actions, object manipulations, and experimental procedures. The benchmark tests procedural reasoning, state tracking, and the ability to translate scientific knowledge into executable actions. To reduce experimental cost, we first retain ScienceWorld instances whose successful reference trajectories can be completed within 30 interaction steps, and obtain 22 task types containing more than five eligible instances. For each retained task type, we sample three instances for training, one for validation, and up to five for testing. If fewer than five eligible test instances remain for a task type, we include all available instances without resampling or duplication. The resulting dataset contains 66 training, 22 validation, and 102 test examples.

\subsection{Baselines}
\label{app:baselines}

\textbf{Handcrafted.}
We use a manually designed ReAct-style agent \citep{yao2023react} with task-level memory as the fixed, non-automated baseline. The agent follows a standard reasoning-and-acting loop and retains relevant interaction history across steps, but its strategy and execution structure are specified by humans and remain unchanged throughout evaluation. 

\textbf{SkillOpt.}
SkillOpt \citep{yang2026skillopt} performs text-level agent optimization by treating a persistent natural-language skill document as the editable state of a frozen task agent. A separate optimizer converts scored trajectories into bounded textual edits, and candidate updates are accepted through held-out validation evaluations. The task model and execution harness remain fixed, so the search space is restricted to the agent's textual strategy representation.

\textbf{AHE.} 
Agentic Harness Engineering (AHE) \citep{lin2026agentic} performs architecture-level optimization over the agent harness. It exposes multiple editable components, including prompts, tools, middleware, skills, memory, and other harness modules, and iteratively modifies these components. Compared with SkillOpt, AHE searches over a broader agent architecture and harness. 

\textbf{Meta-Harness.}
Meta-Harness \citep{lee2026meta} performs end-to-end optimization over the entire model harness. It uses an agentic proposer with filesystem access to the source code, evaluation scores, and execution traces of all previous candidates, which allows arbitrary code-level revisions and organizes cross-round experience as a raw candidate archive. 

\textbf{DGM-H.} 
DGM-Hyperagents (DGM-H) \citep{zhang2026hyperagents} represents the broadest full-code-level automated design baseline. It integrates the task agent and meta agent into a single editable program and allows the resulting hyperagent to modify both its task-solving logic and the mechanism used to generate future improvements. The method searches over arbitrary program logic, memory, and control flow, which provides a substantially less constrained code-level design space.

\begin{table*}[t]
\centering
\caption{Comparison of baseline design scope and cross-round experience organization.}
\label{tab:baseline_scope}
\small
\resizebox{\textwidth}{!}{
\begin{tabular}{llll}
\toprule
Method & Design Level & Editable Scope & Cross-Round Experience\\
\midrule
Handcrafted & Fixed & None & None \\
SkillOpt \citep{yang2026skillopt} & Prompt & Skill document & Accepted/rejected edits and feedback \\
AHE \citep{lin2026agentic} & Harness & Pre-defined components & Records of edits, predictions, and outcomes \\
Meta-Harness \citep{lee2026meta} & Full code & Full agent program & Full candidate archive / raw history \\
DGM-H \citep{zhang2026hyperagents} & Full code & Full agent program & Full candidate archive / raw history \\
ADIAS (Ours) & Full code & Full agent program &  Full candidate archive \& Persistent issue state \\
\bottomrule
\end{tabular}
}
\end{table*}

\subsection{Evaluation Protocol}
\label{app:evaluation-protocol}

We evaluate each method on held-out test tasks after selecting the best agent according to validation performance. All methods use the same task splits, environment interfaces, evaluation budgets, and benchmark-specific evaluators. 

\textbf{Task Performance.}
We retain the native evaluation signal of each benchmark and define the
reported task score as
\begin{equation}
\mathrm{Score}
=
\begin{cases}
\displaystyle
\frac{1}{N}
\sum_{i=1}^{N}
\mathbb{I}[y_i = 1],
&
\text{$\tau$-Bench, ALFWorld, TextCraft},
\\[8pt]
\displaystyle
\frac{1}{N}
\sum_{i=1}^{N}
r_i,
&
\text{WebShop, ScienceWorld},
\end{cases}
\label{eq:task_score}
\end{equation}
where $N$ denotes the number of evaluation tasks, $y_i \in \{0,1\}$ is the binary task-success indicator, and $r_i \in [0,1]$ is the normalized benchmark-specific task score. For $\tau$-Bench, ALFWorld, and TextCraft, task performance is measured by the success rate. For WebShop and ScienceWorld, we report the average native task score, which provides graded credit for partial task progress. The native reward of WebShop measures how well the purchased product satisfies the attributes and constraints specified in the shopping instruction \citep{yao2022webshop}. The score of ScienceWorld captures the degree of progress toward completing the required scientific procedure \citep{wang2022scienceworld}. For presentation, all reported scores are multiplied by 100. We report Pass 1 results by default.

\textbf{Task Interaction Efficiency.}
We measure how effectively a task agent converts environment interactions into
task performance. Let $\bar{L}$ denote the average number of environment interaction steps and $S$ denote the benchmark task score defined above. We define interaction efficiency as
\begin{equation}
    \mathrm{Eff.}
    =
    \frac{Score}{\bar{L}}.
    \label{eq:interaction_efficiency2}
\end{equation}
This performance-per-interaction ratio jointly accounts for task effectiveness and environment interaction cost. We compare efficiency only within the same benchmark, since the semantics and granularity of an environment step differ across domains.

\textbf{Optimization Efficiency.}
Beyond the performance of the final agent, we evaluate how rapidly an automated design method discovers stronger agents under a fixed optimization budget. Let $V_t$ denote the validation score of the agent generated at optimization iteration $t$. We track the best-so-far validation score
\begin{equation}
    B_t
    =
    \max_{1 \leq j \leq t} V_j,
    \label{eq:best_so_far}
\end{equation}
which measures the quality of the agent available after $t$ optimization iterations. A method with a rapidly increasing $B_t$ discovers high-performing designs with fewer search iterations and is therefore more optimization-efficient.

\section{More results}
\label{app:moreresults}

We further examine robustness to execution randomness, optimization variability, finite test-set sampling, and broader task distributions.

\begin{table*}[t]
\centering
\caption{
Multiple runs of evaluation on $\tau$-Bench-Return. Pass@2 counts a task as successful if either rollout succeeds, whereas Pass$^2$ requires both rollouts to succeed. Best and second-best results are highlighted in \highlightbox{best}{green} and \highlightbox{second}{blue}, respectively.
}
\label{tab:evaluation_robustness}
\small
\begin{tabular}{lccccc}
\toprule
Method
& Pass 1
& Pass 2
& Pass@2
& Pass$^2$
& Avg. $\pm$ Std. \\
\midrule

Handcrafted
& 75.0
& 68.8
& 87.5
& 56.3
& 71.9 $\pm$ 4.4 \\\

SkillOpt
& 12.5
& 6.3
& 15.6
& 3.1
& 9.4 $\pm$ 4.4 \\\

Meta-Harness
& 43.8
& 43.8
& 53.1
& 34.4
& 43.8 $\pm$ 0.0 \\\

AHE
& 56.2
& 37.5
& 68.8
& 25.0
& 46.9 $\pm$ 13.2 \\\

DGM-H
& 59.4
& 43.8
& 65.6
& 37.5
& 51.6 $\pm$ 11.0 \\\

ADIAS (Ours)
& \cellcolor{best}81.3
& \cellcolor{best}81.3
& \cellcolor{best}90.6
& \cellcolor{best}71.9
& \cellcolor{best}81.3 $\pm$ 0.0\\

\midrule
\multicolumn{6}{l}{\textit{Ablations}} \\

\quad w/o External Prior
& 75.0
& \cellcolor{second}75.0
& \cellcolor{best}90.6
& 59.4
& \cellcolor{second}75.0 $\pm$ 0.0\\

\quad w/o Round-Level Diagnosis
& \cellcolor{second}78.1
& 71.9
& \cellcolor{second}87.5
& \cellcolor{second}62.5
& \cellcolor{second}75.0 $\pm$ 4.4\\

\quad w/ Archive-Wide Synthesis
& 65.6
& 62.6
& 78.1
& 50.0
& 64.1 $\pm$ 2.1 \\

\quad w/ Best-Candidate Revision
& 71.9
& 71.9
& 81.3
& \cellcolor{second}62.5
& 71.9 $\pm$ 0.0 \\

\quad w/ Latest-Candidate Continuation
& 62.5
& 62.5
& 75.0
& 50.0
& 62.5 $\pm$ 0.0 \\

\bottomrule
\end{tabular}

\end{table*}

\subsection{Evaluation Robustness}
\label{app:evaluation_robustness}

We examine whether the performance of a fixed selected agent is robust to execution randomness by repeating its evaluation on the same held-out test instances. The main experiments report the result of the first evaluation pass for a consistent single-rollout comparison across benchmarks. We conduct a second independent rollout on $\tau$-Bench to examine whether the reported performance is robust to execution randomness. Both passes use the same agents, task instances, and evaluation protocol.

For task $i$, let $y_i^{(j)}\in{\{0,1\}}$ denote whether the $j$-th rollout succeeds. We report two complementary repeated-execution metrics. Pass@${k}$ measures whether a task succeeds in at least one of the $k$ rollouts:
\begin{equation}
\mathrm{Pass@}k
=
\frac{1}{N}
\sum_{i=1}^{N}
\left(
1-
\prod_{j=1}^{k}
\left(1-y_i^{(j)}\right)
\right).
\label{eq:pass_at_k}
\end{equation}
In contrast, Pass$^{k}$ measures consistent success across all $k$ rollouts \citep{yao2025tau}:
\begin{equation}
\mathrm{Pass}^{k}
=
\frac{1}{N}
\sum_{i=1}^{N}
\prod_{j=1}^{k} y_i^{(j)}.
\label{eq:pass_power_k}
\end{equation}
Pass@${k}$ therefore captures whether the agent can solve a task in multiple runs, whereas Pass$^{k}$ provides a stricter measure of whether it solves the task reliably. We additionally report the arithmetic mean of the two individual pass scores. The same score of two passes merely indicates that the success rates of the two aggregations are the same; it does not imply that the two aggregations resolved exactly the same set of tasks. The consistency of the specific tasks is more accurately reflected by Pass$^{k}$.

Overall, the repeated-evaluation results preserve the conclusions of the main experiments and show that ADIAS illustrates robust execution performance rather than a favorable outcome from a single rollout. As shown in Table~\ref{tab:evaluation_robustness}, ADIAS obtains identical success rates of 81.3 across the two independent evaluation passes, exceeding the strongest baseline average of 71.9 by 9.4 points. More importantly, ADIAS achieves a Pass$^2$ score of 71.9, outperforming the strongest non-ADIAS baseline by 15.6 points. This indicates that its advantage is not driven solely by tasks that succeed under favorable sampling; a substantially larger fraction of tasks are solved consistently across both executions.
The ablations further support the importance of the proposed design. Removing either the external prior or round-level diagnosis reduces the two-pass average to 75.0 and lowers Pass$^2$ to 59.4 and 62.5, respectively. Replacing the persistent issue-centric state with candidate-based revision policies produces larger reductions in both average performance and consistent success. 

\begin{table}[t]
\centering
\caption{
Optimization-run robustness on ALFWorld across three independent complete optimization runs. Avg. $\pm$ Std. reports the mean and standard deviation of test scores across runs.
}
\label{tab}
\small
\begin{tabular}{lccc}
\toprule
Method
& Avg. $\pm$ Std. $\uparrow$
& Pass@3 $\uparrow$
& Pass$^3$ $\uparrow$ \\
\midrule
SkillOpt
& 0.0 $\pm$ 0.0
& 0.0
& 0.0 \\
Meta-Harness
& 54.7 $\pm$ 19.2
& 89.6
& 23.1 \\
AHE
& 81.1 $\pm$ 2.4
& 88.8
& 68.7 \\
DGM-H
& 71.6 $\pm$ 3.3
& 93.3
& 46.3 \\
ADIAS (Ours)
& \textbf{93.3 $\pm$ 1.3}
& \textbf{98.5}
& \textbf{86.6} \\
\bottomrule
\end{tabular}
\end{table}
\subsection{Optimization-Run Robustness}
We examine whether the complete optimization procedure consistently discovers strong agent designs by repeating the full optimization process in three independent runs on ALFWorld. Figure~\ref{fig:optimization_progress} reports the mean best-so-far validation trajectories, while Table~\ref{tab} summarizes the held-out test performance of the three independently optimized agents. Avg. $\pm$ Std. measures variation in overall test performance across optimization runs. ADIAS achieves both the highest average test performance and the lowest variability among the competitive methods, reaching 93.3 $\pm$ 1.3 across three independent optimization runs. It also attains 98.5\% Pass@3 and 86.6\% Pass$^3$, which shows that its optimization procedure not only discovers agents with broad task coverage, but also reproduces successful behavior on most test tasks across independently optimized agents.

\begin{table}[t]
\centering
\caption{
Generalization from the \emph{Clean-and-Place} task family to the broader ALFWorld task distribution. All methods are optimized only on \emph{Clean-and-Place} training and validation examples. 
}
\label{tab:alfworld_generalization}
\small
\begin{tabular}{lcccc}
\toprule
& \multicolumn{2}{c}{Clean-and-Place Test}
& \multicolumn{2}{c}{Official Test Set} \\
\cmidrule(lr){2-3}
\cmidrule(lr){4-5}
Method
& Score $\uparrow$
& Eff. $\uparrow$
& Score $\uparrow$
& Eff. $\uparrow$ \\
\midrule
Handcrafted
& 76.7 & 4.26
& 70.9 & 3.60 \\

SkillOpt
& 5.0 & 0.17
& 0.0 & 0.00 \\

Meta-Harness
& 83.3 & 6.08
& 59.7 & 3.32 \\

AHE
& 60.0 & 3.49
& 78.4 & 4.48 \\

DGM-H
& 83.3 & 5.05
& 69.4 & 3.42 \\

ADIAS (Ours)
& \textbf{86.7} & \textbf{8.03}
& \textbf{94.0} & \textbf{8.95} \\
\bottomrule
\end{tabular}
\end{table}

\subsection{Task-Family Generalization}
\label{app:task_generalization}

We examine whether agents optimized on a single ALFWorld task family generalize to a broader task distribution. All methods are optimized using 153 training and 15 validation examples from \emph{Clean-and-Place}. We evaluate the resulting agents on two test sets: 30 held-out \emph{Clean-and-Place} examples for in-distribution evaluation, and the official ALFWorld test set of 134 examples spanning multiple task families for cross-task evaluation.

ADIAS achieves the highest success rate on both the held-out \emph{Clean-and-Place} set and the broader official test set, while also requiring the fewest interaction steps. Notably, although optimization is performed exclusively on \emph{Clean-and-Place}, ADIAS reaches 94.0\% success on the official multi-task test distribution, compared with 78.4\% for the strongest baseline. These results indicate that the optimized harness does not merely specialize to the task family seen during optimization, but transfers effectively to other ALFWorld task types.

\subsection{Expanded-Distribution Evaluation}
\label{app:full_retail}

We examine whether the observed advantage persists when optimization and evaluation are expanded from the main task subset to a broader task distribution. The main experiments use the return-intent subset of $\tau$-Bench to reduce the substantial costs. To verify that the observed gains are not specific to this task family, we further optimize and evaluate all methods on the complete $\tau$-Bench retail distribution, which contains 500 training, 20 validation, and 115 test tasks spanning a broader range of customer-service requests and interaction patterns. Under this expanded setting, each method is optimized for 15 iterations using 20 training episodes per iteration.

As shown in Table~\ref{tab:taubench_full}, ADIAS remains the strongest method when optimization is scaled to the complete retail distribution. It achieves a Pass-1 success rate of 81.7, outperforming the strongest baseline by 13.0 percentage points. The improvement is not obtained at the expense of interaction efficiency: ADIAS reaches an efficiency of 6.24, compared with 5.13 for the strongest baseline.

The advantage also persists across repeated executions. ADIAS obtains an average success rate of 80.4 $\pm$ 1.8, exceeding the strongest baseline average of 67.8 $\pm$ 1.2 by 12.6 points. Its Pass$^2$ score reaches 68.7, outperforming the best baseline by 14.8 points, while its Pass@2 score of 92.2 indicates broad task coverage across the two executions. These results show that ADIAS improves both task performance and execution consistency on the broader retail distribution.

\begin{table}[t]
\centering
\caption{
Results on the complete Tau-Bench retail distribution. Pass@2 counts a task as successful if either evaluation rollout succeeds, whereas Pass$^2$ requires both rollouts to succeed.  Best and second-best results are highlighted in \highlightbox{best}{green} and \highlightbox{second}{blue}, respectively.
}
\label{tab:taubench_full}
\small
\begin{tabular}{lcccccc}
\toprule
Method
& Pass 1 $\uparrow$
& Eff. $\uparrow$
& Pass 2 $\uparrow$
& Pass@2 $\uparrow$
& Pass$^2$ $\uparrow$
& Avg. $\pm$ Std. $\uparrow$ \\
\midrule

Handcrafted
& \cellcolor{second}68.7
& \cellcolor{second}5.13
& \cellcolor{second}67.0
& \cellcolor{second}81.7
& \cellcolor{second}53.9
& \cellcolor{second}67.8 $\pm$ 1.2 \\

SkillOpt
& 6.1
& 0.23
& 5.2
& 7.8
& 3.5
& 5.7 $\pm$ 0.6 \\

Meta-Harness
& 58.1
& 4.12
& 58.1
& 69.5
& 46.7
& 58.1 $\pm$ 0.0 \\

AHE
& 67.8
& 4.13
& 64.3
& 80.9
& 51.3
& 66.1 $\pm$ 2.5 \\

DGM-H
& 63.5
& 3.94
& 59.1
& 77.4
& 45.2
& 61.3 $\pm$ 3.1 \\

ADIAS (Ours)
& \cellcolor{best}81.7
& \cellcolor{best}6.24
& \cellcolor{best}79.1
& \cellcolor{best}92.2
& \cellcolor{best}68.7
& \cellcolor{best}80.4 $\pm$ 1.8 \\

\bottomrule
\end{tabular}

\end{table}

\begin{table}[t]
\centering
\caption{
Test-set uncertainty across three benchmarks. Each cell reports the task score followed by its 95\% confidence interval.
The strongest baseline is Handcrafted for Tau-Bench Retail, AHE for ALFWorld, and DGM-H for TextCraft. The difference row reports ADIAS minus the strongest baseline with paired 95\% confidence intervals.
}
\label{tab:test_ci}
\small
\begin{tabular}{lccc}
\toprule
Method
& Tau-Bench Retail
& ALFWorld
& TextCraft
 \\
\midrule

Strongest Baseline
& 68.70 [59.73, 76.45]
& 78.36 [70.65, 84.49]
& 76.00 [66.77, 83.31]
 \\

ADIAS (Ours)
& \textbf{81.74 [73.69, 87.73]}
& \textbf{94.03 [88.66, 96.94]}
& \textbf{91.00 [83.77, 95.19]}
 \\

$\Delta$ (Ours $-$ Baseline)
& \textbf{+13.04 [4.35, 21.74]}
& \textbf{+15.67 [8.96, 23.13]}
& \textbf{+15.00 [7.00, 24.00]}
 \\

\bottomrule
\end{tabular}
\end{table} 
\subsection{Test-Set Uncertainty}
The main results report the score estimates on finite held-out test sets. We additionally quantify uncertainty due to test-set sampling using nonparametric task-level bootstrap resampling. For each benchmark, we compare ADIAS with the strongest competing method in Table \ref{tab:test_ci}. We resample test instances with replacement 10,000 times and use identical sampled indices for both methods. We report 95\% confidence intervals for each method's task performance and for their paired performance difference. Across three evaluated benchmarks, the paired 95\% confidence intervals remain entirely above zero, indicating that ADIAS's advantage over the strongest competing method is robust to finite test-set sampling. 

\section{Additional Analysis and Findings}
\subsection{Trade-off between Expressiveness and Searchability}
The automated agent design baselines exhibit different behaviors as the editable agent-design space expands from text/prompt to harness/architecture components and full code level. 

SkillOpt operates at the text level by optimizing a persistent skill document while leaving the agent structure fixed. This restricted design space enables controlled updates, but prevents the optimizer from introducing new modules when the task requires capabilities absent from the initial agent. For example, on ALFWorld, effective long-horizon interaction requires persistent state tracking and memory mechanisms. As SkillOpt cannot alter the agent architecture to introduce such mechanisms, its performance is significantly lower than that of the other methods.

AHE expands the editable space to the agent harness and can modify multiple components, such as prompts, tools, skills, and memory. This broader scope enables useful improvements across different environments, and AHE achieves non-trivial gains on all five benchmarks. However, its search remains constrained by a predefined set of harness components and interfaces. As a result, the performance of AHE generally falls to the middle of the performance range. 

DGM-H removes this structural restriction by allowing full-code-level self-modification. Its larger design space provides a higher performance ceiling, as illustrated by its strong result on ALFWorld. However, unrestricted code search introduces a substantially more difficult optimization problem. Under a limited evaluation budget, useful agent designs may be extremely sparse in the space of arbitrary program modifications. Thus, DGM-H fails to consistently outperform other baselines.

ADIAS resolves this trade-off between \emph{expressiveness} and
\emph{searchability}. Similar to DGM-H, ADIAS retains a full-code design space and can introduce new agent mechanisms when required by the task. Unlike unconstrained code-level evolution, however, ADIAS narrows the effective search space with task priors that guide the designer toward mechanisms likely to be useful for the target environment. In addition, the \emph{diagnostic agent} performs fine-grained failure analysis and precisely attributes observed errors to actionable causes. Finally, a global \emph{Issue Manager} aggregates recurring patterns across episodes and iterations to prioritize broadly useful modifications and improve optimization efficiency. Overall, ADIAS preserves a high-capacity search space while using task priors to guide exploration, precise diagnosis to localize failures, and global issue state to drive efficient improvement.

\subsection{Optimization of the Optimization Process Itself}
\label{app:optimizer_evolution}

Because ADIAS optimizes the full agent harness, its editable space is not restricted to the task agent. Components involved in diagnosing failures and planning subsequent revisions may also be modified, which allows the optimization process itself to become part of the search space. 
We observe several such modifications during optimization. The Issue Manager is revised to normalize the revision plan into a fixed schema containing the target failure mode, selected parent generation, and revision rationale. The code improver is also revised to simplify how the target failure mode is extracted from the revision plan. These edits indicate that ADIAS can identify and revise interfaces within the optimization machinery, rather than modifying only task-execution behavior.

\begin{table}[t]
\centering
\caption{
Meta-level token consumption and held-out test performance on Tau-Bench and ALFWorld.
Meta tokens include LLM calls used by the optimization procedure. All token values are in millions.
}
\label{tab:token_cost}
\small
\begin{tabular}{lcccc}
\toprule
& \multicolumn{2}{c}{$\tau$-Bench-Return}
& \multicolumn{2}{c}{ALFWorld} \\
\cmidrule(lr){2-3}
\cmidrule(lr){4-5}
Method
& Meta Tokens $\downarrow$
& Score $\uparrow$
& Meta Tokens $\downarrow$
& Score $\uparrow$ \\
\midrule

SkillOpt
&1.0 & 12.5
& 1.0 & 0.0 \\

Meta-Harness
& 3.1 & 43.8
& 4.0 & 59.7 \\

AHE
& 4.4 & 56.2
& 3.9 & 78.4 \\

DGM-H
& 4.7 & 59.4
& 3.2 & 69.4 \\

ADIAS (Ours)
& 8.1 & 81.3
& 6.3 & 94.0 \\

\midrule
\multicolumn{5}{l}{\textit{Ablations}} \\

\quad w/o External Prior
& 12.5 & 75.0
& 9.1 & 82.8 \\

\quad w/o Round-Level Diagnosis
& 2.9 & 78.1
& 4.5 & 63.4 \\

\quad w/ Archive-Wide Synthesis
& 7.7 & 65.6
& 5.6 & 60.4 \\

\quad w/ Best-Candidate Revision
& 13.1 & 71.9
& 10.7 & 75.4 \\

\quad w/ Latest-Candidate Continuation
& 8.2 & 62.5
& 12.6 & 41.0 \\

\bottomrule
\end{tabular}
\end{table}

\subsection{Token Consumption}
\label{app:token_cost}

We report \emph{meta-level token consumption} in Table~\ref{tab:token_cost} together with held-out test performance on $\tau$-Bench and ALFWorld. It measures the LLM calls used by the optimization procedure itself. We do not use total token consumption for direct comparison because task-rollout cost is also affected by the behavior of the generated agents, and therefore does not isolate the computational overhead of the optimization method. 
ADIAS uses more meta-level tokens than other automated baselines, which reflects the additional reasoning required for diagnosis and persistent state maintenance. However, the controlled variants indicate that the performance gains cannot be explained simply by greater meta-level model usage. On $\tau$-Bench, Archive-Wide Synthesis and Latest-Candidate Continuation consume nearly the same number of tokens as ADIAS (7.7M and 8.2M vs.\ 8.1M), yet their scores drop from 81.3 to 65.6 and 62.5, respectively. The pattern is even clearer on ALFWorld: Archive-Wide Synthesis uses 5.6M tokens but scores only 60.4, while Best-Candidate Revision and Latest-Candidate Continuation consume substantially more tokens than ADIAS (10.7M and 12.6M) while reaching only 75.4 and 41.0. Removing round-level diagnosis reduces meta-level cost, but also reduces performance.

\subsection{Limitations and Future Work}

First, issue-centric optimization relies on sufficiently accurate trajectory diagnosis and issue consolidation. In this work, diagnosis is treated as an intermediate component rather than a prediction task; its effectiveness is therefore evaluated through downstream optimization performance and qualitative case analysis with human inspection. The strong gains across benchmarks and controlled ablations suggest that the resulting diagnoses are sufficiently informative for guiding optimization, although a large-scale annotation study of diagnosis accuracy is beyond the scope of this work. Future work could further improve diagnosis and issue association, or develop explicit benchmarks for evaluating them. Second, our evaluation focuses on text-based interactive environments. While the experiments cover multiple benchmarks, backbone models, repeated optimization runs, and broader task distributions, extending issue-centric optimization to substantially longer-horizon tasks and multimodal agent environments is a promising direction for future work. 


\end{document}